%% file: main.tex
\documentclass[10pt,twocolumn,letterpaper]{article}
\makeatletter
\@namedef{ver@everyshi.sty}{}
\makeatother
\usepackage{iccv}
\usepackage{times}
\usepackage{epsfig}
\usepackage{graphicx}

\usepackage{amsmath}
\usepackage{dsfont}
\usepackage{amssymb}
\usepackage{pifont} % http://ctan.org/pkg/pifont
\newcommand{\xmark}{\ding{55}}%
\newcommand{\cmark}{\ding{51}}%
\usepackage[table,dvipsnames]{xcolor}
\usepackage{tcolorbox}
\usepackage{adjustbox}

\usepackage{tabularx}
\usepackage{makecell}
\usepackage{multirow}
\usepackage{enumitem}  % for leftmargin

\usepackage[accsupp]{axessibility}  % Improves PDF readability for those with disabilities.

\newcommand{\red}[1]{\textcolor{red}{#1}}
\newcommand{\blueD}[1]{\textcolor{blue}{#1}}

\newcommand{\blue}[1]{\textcolor{cyan}{#1}}

\newcommand{\inlineimg}[1]{\raisebox{-0.2\baselineskip}{\includegraphics[height=0.95\baselineskip]{#1.png}}}

\newcommand{\etv}{EgoTV\xspace}
\newcommand{\nsg}{NSG\xspace}
\newcommand{\nt}{Novel Tasks\xspace}
\newcommand{\nst}{Novel Steps\xspace}
\newcommand{\nc}{Novel Scenes\xspace}

\newcommand{\code}[1]{\textcolor{darkgray}{\texttt{#1}}}
\newcommand{\query}[1]{\textcolor{blue}{\texttt{#1}}}
\newcommand{\answer}[1]{\textcolor{Green}{\texttt{#1}}}

\usepackage[misc]{ifsym}

\usepackage[breaklinks=true,bookmarks=false]{hyperref}

\iccvfinalcopy % *** Uncomment this line for the final submission

\def\iccvPaperID{1945} % *** Enter the ICCV Paper ID here

\ificcvfinal\fi

\begin{document}

%%%%%%%%% TITLE
\title{\etv \inlineimg{figures/TV}: Egocentric Task Verification \\from Natural Language Task Descriptions}
% \title{\etv \emoji(tv): Egocentric Task Verification \\from Natural Language Task Descriptions}
% \and
% \author{%
%     Rishi Hazra$^1$  \quad
%     Brian Chen $^2$  \quad
%     Akshara Rai$^2$  \quad 
%     Nitin Kamra $^2$  \quad
%     Ruta Desai$^2$   \\
%     %\vspace{1mm} \\
%     \small{$^1$Örebro University, $^2$Meta}  \\
%     \small{
%     \texttt{rishi.hazra@oru.se}, \texttt{\{bc2754,nitinkamra,akshararai,rutadesai\}@meta.com}}   
% }

% \author{Rishi Hazra\thanks{Work done while interning at Meta Reality Labs Research.}\\
% Örebro University\\
% \vspace{-0.3cm}
% \and
% Brian Chen\\
% Meta Reality Labs Research\\
% \vspace{-0.3cm}
% \and
% Akshara Rai\\
% Meta AI Research\\
% \vspace{-0.3cm}
% \and
% Nitin Kamra\\
% Meta Reality Labs Research
% \and
% Ruta Desai \Letter\\
% Meta Reality Labs Research\\ 
% \vspace{-1cm}
% \and
% \vspace{-1cm}
% {\tt\small rishi.hazra@oru.se}, {\tt\small{\{bc2754,nitinkamra,akshararai,rutadesai\}@meta.com}}\\
% \vspace{0.7cm}
% \and
% {\tt\href{https://rishihazra.github.io/EgoTV}{\textcolor{magenta}{EgoTV.github.io}}}
% }
\author{
Rishi Hazra\textsuperscript{1}\thanks{Work done while interning at Meta.}, \quad
Brian Chen\textsuperscript{2}, \quad
Akshara Rai\textsuperscript{2}, \quad
Nitin Kamra\textsuperscript{2}, \quad
Ruta Desai\textsuperscript{2}\Letter \quad \\
\textsuperscript{1}Örebro University \quad
\textsuperscript{2}Meta\\
{\tt\normalsize rishi.hazra@oru.se}, {\tt\small{\{bc2754,nitinkamra,akshararai,rutadesai\}@meta.com}} \\
{\tt\normalsize \href{https://rishihazra.github.io/EgoTV}{\textcolor{magenta}{https://rishihazra.github.io/EgoTV}}}
}

\maketitle
\begin{figure*}[t]
\centering
    \includegraphics[width=\linewidth]{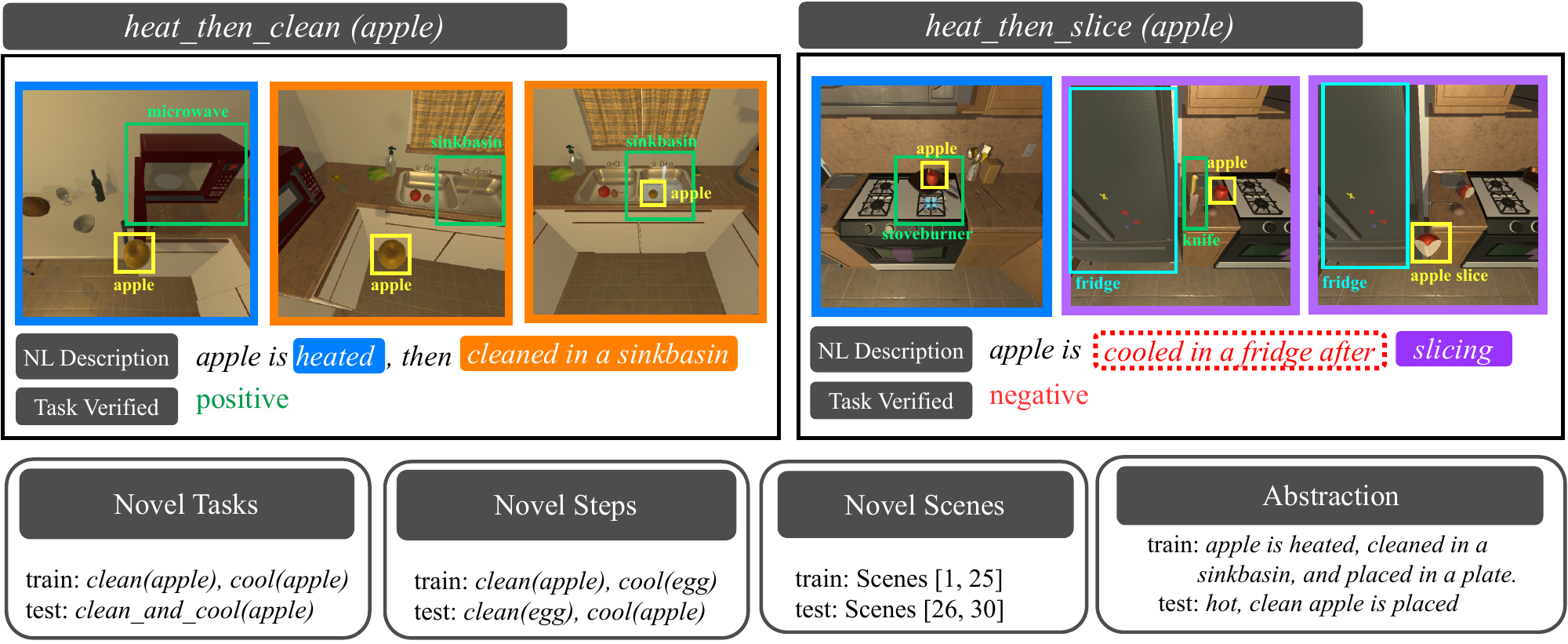}
    \caption{\textbf{\etv benchmark.} A positive example [Left] and a negative example [Right] from the train set along with illustrative examples from the test splits [Bottom] of \etv are shown. The test splits are focused on generalization to novel compositions of tasks, unseen sub-tasks or steps and scenes, and abstraction in NL task descriptions. The bounding boxes are solely for demonstration purposes and are not used during training/inference.}
    
    % are (i) Novel Tasks: unseen compositions of sub-tasks, (ii) Novel Steps: unseen sub-task and target object compositions, and (iii) Novel Scenes: seen tasks in unseen scenes. Additionally, we use the Abstraction split to measure the robustness of the models on abstract task descriptions. Note, that the bounding boxes are solely for demonstration purposes and are not used during training/inference. \aks{I think that the generalization row is a little unclear. Can you remove the train generalization examples, and make all examples a composition of the tasks in the first row? Also can you add a NL description of the generalization task? Finally do we need the underscore ``heat\_then\_clean(apple)", or can we just write ``heat then clean the apple"? In the method section we can describe that we convert the NL description to this template before passing to NSG.}}
    \label{figure:dataset}
\end{figure*}

\ificcvfinal\thispagestyle{empty}\fi

%%%%%%%%%%%%%%%%%%%%%%%%%%%%%%%%%5

%%%%%%%%% ABSTRACT

\input{sections_CRV/0_Abstract}
\input{sections_CRV/1_Introduction}
\input{sections_CRV/2_related_work}
\input{sections_CRV/3_Dataset}
\input{sections_CRV/3.5_CTV}
\input{sections_CRV/4_NSG_for_EgoTV}
\input{sections_CRV/5_Experiments}
\input{sections_CRV/6_Conclusion}
\input{sections_CRV/7_Acknowledgement}
%\section{Discussion}

{\small
\bibliographystyle{ieee_fullname}
\bibliography{egbib}
}

\input{sections_CRV/sup}
\end{document}

%% file: sections_CRV/0_Abstract.tex
% \begin{abstract}
%  Towards next generation of assistants, our goal is 
%  \rd{TODO: abstract}
%  to make progress towards egocentric agents capable of tracking and verifying everyday tasks specified in natural language. Such agents should be able to reason about various ways of doing multi-step tasks, decompose the tasks into relevant actions, state changes, object interactions, and any necessary causal relationships. To that end, we propose a benchmark and dataset called Egocentric Task Verification (EgoTV) using a photo-realistic, embodied simulator called AI2-THOR. EgoTV contains multi-step tasks with ordering constraints and abstracted task descriptions with the omission of low-level task details. We also introduce a novel Neuro-Symbolic Grounding (NSG) approach to enable the core reasoning capability for EgoTV. We demonstrate NSG's capability on our synthetic EgoTV dataset and a real-world dataset derived from Crosstask. Our contributions include EgoTV, NSG, and the release of both datasets and models for future egocentric task tracking and verification research.
% \end{abstract}

\begin{abstract}
To enable progress towards egocentric agents capable of understanding everyday tasks specified in natural language, we propose a benchmark and a synthetic dataset called Egocentric Task Verification (\etv). The goal in \etv is to verify the execution of tasks from egocentric videos based on the natural language description of these tasks. \etv contains pairs of videos and their task descriptions for multi-step tasks -- these tasks contain multiple sub-task decompositions, state changes, object interactions, and sub-task ordering constraints. In addition, \etv also provides abstracted task descriptions that contain only partial details about ways to accomplish a task. Consequently, \etv requires causal, temporal, and compositional reasoning of video and language modalities, which is missing in existing datasets. We also find that existing vision-language models struggle at such all round reasoning needed for task verification in \etv. Inspired by the needs of \etv, we propose a novel Neuro-Symbolic Grounding (NSG) approach that leverages symbolic representations to capture the compositional and temporal structure of tasks. We demonstrate NSG's capability towards task tracking and verification on our \etv dataset and a real-world dataset derived from CrossTask~\cite{cross_task} (CTV). We open-source the \etv and CTV datasets and the NSG model for future research on egocentric assistive agents. 
% \etv's source code is publicly available.
% \footnote{ \url{https://github.com/facebookresearch/EgoTV}}
\end{abstract}

%% file: sections_CRV/1_Introduction.tex
% \vspace{-10pt}
%%%%%%%%% BODY TEXT

\section{Introduction}
\label{section:introduction}

Inspired by recent progress in visual systems~\cite{MagicLeap, ungureanu2020hololens}, we consider an assistive egocentric agent capable of reasoning about daily activities. When invoked via natural language commands, for e.g., while baking a cake, the agent understands the steps involved in baking, tracks progress through the various stages of the task, detects and proactively prevents mistakes by making suggestions. Such a virtual agent~\cite{virtual-agent} would empower users to learn new skills and accomplish tasks efficiently.

Developing this egocentric agent capable of tracking and verifying everyday tasks based on their natural language specification is challenging for multiple reasons. First, such an agent must reason about various ways of doing a \emph{multi-step} task specified in natural language. This entails decomposing the task into relevant actions, state changes, object interactions as well as any necessary causal and temporal relationships between these entities. Secondly, the agent must ground these entities in egocentric observations to track progress and detect mistakes. Lastly, to truly be useful, such an agent must support tracking and verification for a combination of tasks and, ideally, even unseen tasks. These three challenges -- causal and temporal reasoning about task structure from natural language, visual grounding of sub-tasks, and compositional generalization -- form the core goals of our work.

As our first contribution, we propose a benchmark -- \emph{\textbf{Ego}centric \textbf{T}ask \textbf{V}erification} (\etv \inlineimg{figures/TV}) -- and a corresponding dataset in the AI2-THOR~\cite{ai2thor} simulator. % \emoji{tv}
Given a natural language (NL) task description and a corresponding egocentric video of an agent, the goal of \etv is to verify whether the task was successfully completed in the video or not.
\etv contains multi-step tasks with \emph{ordering} constraints on the steps and \emph{abstracted} NL task descriptions with omitted low-level task details inspired by the needs of real-world assistants. We also provide splits of the dataset focused on different generalization aspects, e.g., unseen visual contexts, compositions of steps, and tasks (see Figure~\ref{figure:dataset}). Consequently, \etv dataset provides the fine-grained control necessary for rigorous testing and refinement of task reasoning models, which is often missing in real-world datasets~\cite{ego_4d, epic_kitchens}. Yet, \etv mirrors the real world by leveraging visual photo-realism and task diversity.

Our second contribution is a novel approach for order-aware visual grounding~--~\emph{\textbf{N}euro-\textbf{S}ymbolic \textbf{G}rounding} (NSG), capable of compositional reasoning and generalizing to unseen tasks owing to its ability to leverage abstract NL descriptions along with compositional and temporal structure of tasks (task decomposition, ordering).~In contrast, state-of-the-art vision-language models~\cite{coca,clip,videoclip,clip_hitchiker} struggle to ground NL descriptions in egocentric videos, and do not generalize to unseen tasks.~NSG outperforms these models by~$\mathbf{33.8}\%$~on compositional generalization and~$\mathbf{32.8}\%$~on abstractly described task verification. Finally, to evaluate \nsg on real-world data, we instantiate \etv on the CrossTask~\cite{cross_task} instructional video dataset. We find that it also outperforms state-of-the-art models at task verification on CrossTask. We hope that the \etv~benchmark and dataset will enable future research on egocentric agents capable of aiding in everyday tasks.
%%%%%%%%%%%%%%%%%%%%%%%%%%%%%%%%%%%%%%%%%%%%%%%%%%%%%%%%%%%

%% file: sections_CRV/2_related_work.tex
\section{Related Work}
\label{section:related_works}

\noindent \textbf{Video-based Task Understanding.} 
Understanding tasks from videos has been a long-standing theme in vision research with focus on recognizing activities~\cite{charades_dataset, epic_kitchens}, human-object interactions~\cite{action_genome, ego_4d}, and object state changes~\cite{change_it, fathi2013modeling} using egocentric or exocentric videos. But apart from recognizing actions, objects, and state changes, task verification also requires understanding temporal orderings between them. Our work is, therefore, closer to research on understanding instructional tasks~\cite{cross_task, tang2019coin}, which require reasoning about multiple, ordered steps. Prior works focus on either learning the order of steps~\cite{bansal2022my, lin2022learning, huang2019neural, mao2023action} or use step-ordering as a supervisory signal for learning step-representations or step-segmentation~\cite{cross_task, shen2021learning}. Instead, we are focused on video-based order verification of steps described in NL, akin to~\cite{qian2022svip}.

\noindent \textbf{Temporal Video Grounding.} Our \etv benchmark is also closely related to the problem of Temporal Video Grounding (TVG)~\cite{mexaction2,human_activity_understanding,regneri2013grounding,change_it}.
However, prior work on TVG predominantly focuses on localizing a single action in the video~\cite{MAD_dataset,jiang2014thumos}.
In contrast, \etv requires localizing multiple actions, wherein actions could have partial ordering, i.e., actions could have more than one valid ordering amongst them. 

\input{tables/Tab_1_abridged}

\noindent \textbf{Vision-Language Benchmarks.} 
Various benchmark tasks have been proposed for enabling models that can reason across video and language modalities (see Table~\ref{table:list_of_datasets_abridged}). Examples include video question answering~\cite{clevrer,next_qa_dataset,agqa_dataset,activity_net_dataset,star_situated_reasoning,tvqa_dataset,movie_qa,cater_dataset}, video-based entailment~\cite{violin_dataset}, and embodied task completion~\cite{ALFRED20,teach_alexa,behavior_benchmark}. However, these benchmarks focus on individual specific aspects of multimodal reasoning, e.g., compositional reasoning (AGQA~\cite{agqa_dataset}, ActivityNet-QA~\cite{activity_net_dataset}, TVQA~\cite{tvqa_dataset}, and CATER~\cite{cater_dataset}) or causal reasoning (NExT-QA~\cite{next_qa_dataset}, CoPhy~\cite{cophy_dataset}, Causal-VidQA~\cite{causal_vid_qa}, EgoTaskQA\cite{jia2022egotaskqa}, and VIOLIN~\cite{violin_dataset}). In comparison, \etv focuses on both causal and compositional reasoning and further requires visual grounding of both objects and actions from text, similar to STAR~\cite{star_situated_reasoning} and CLEVRER~\cite{clevrer}, albeit in egocentric settings. Unlike embodied task completion benchmarks whose objective is to develop robotic agents that can \emph{perform everyday tasks} through task-planning (ALFRED~\cite{ALFRED20}, TEACh~\cite{teach_alexa}) and control (Behavior~\cite{behavior_benchmark}), EgoTV benchmark's objective is to develop virtual agents that can \emph{track and verify everyday tasks} performed by humans. Akin to NLP \emph{Entailment} problem~\cite{pascal_text_entailment,snli_ve_dataset}, it can also be viewed as a video-based entailment problem -- where a given ``premise" (video) is validated by a ``hypothesis" (task description).
% Unlike embodied instruction following (ALFRED~\cite{ALFRED20}, TEACh~\cite{teach_alexa}), which requires interpreting grounded task instructions for task execution, \etv focuses on task verification in egocentric videos from instruction-like descriptions. Behavior benchmark's~\cite{behavior_benchmark} objective is to develop robotic agents that can \emph{perform everyday tasks}, evaluated using standard reinforcement learning baselines. Instead, EgoTV benchmark's objective is to develop virtual agents that can \emph{track and verify everyday tasks} performed by humans. 

\noindent \textbf{Vision-Language Models.}
Vision-Language Models (VLMs)~\cite{clip,videoclip,luo2022clip4clip,blip,coca} pre-trained on large-scale image-text or video-language narration pairs have demonstrated enhanced performance on certain compositional~\cite{li2020hero} and causal~\cite{change_it} tasks. However, they generally struggle to handle compositionality and order sensitivity~\cite{VLMbag-of-words,winoground}. Instead, NSG explicitly targets order awareness and compositionality for generalization in task verification using neuro-symbolic reasoning.

\noindent \textbf{Neuro-symbolic Models.} Neuro-symbolic models combine feature extraction through deep learning with symbolic reasoning~\cite{nesy-survey, star_situated_reasoning} to capture compositional substructures. These models either reason on static images to recognize object attributes and relations (NS-CL~\cite{nscl}, NS-VQA~\cite{nsvqa}, CLOSURE~\cite{closure}, and $\nabla-$FOL~\cite{del-fol}), or on videos to recognize spatio-temporal and causal relations (NS-DR~\cite{clevrer} and DCL~\cite{dcl}). We extend this to tracking multi-step actions.

%%%%%%%%%%%%%%%%%%%%%%%%%%%%%%%%%%%%%%%%%%%%%%%%%%%%%%%%%%%%%%%%%

%% file: tables/Tab_1_abridged.tex
\begin{table*}[t]
\centering
% \footnotesize
%\footnotesize{
\begin{tabular}{lcccccc}
% \hline
& \multicolumn{3}{c}{\textbf{-------------- Reasoning --------------}} & \multicolumn{3}{c}{\textbf{------  Dataset Characteristics ------}} \\
 & compositional & causal & temporal & egocentric & \begin{tabular}[c]{@{}c@{}}real-\\ world\end{tabular} & \begin{tabular}[c]{@{}c@{}}diagnostic\\ tools\end{tabular}\\
\hline
% CLEVR~\cite{clevr_dataset} & \multicolumn{3}{l}{\qquad \cmark $\kern 2.3pc$ \xmark $\kern 2.7pc$ \xmark} & \multicolumn{5}{l}{\qquad \cmark $\kern 2.7pc$ \xmark $\kern 2.8pc$ \xmark $\kern 2.8pc$ \xmark $\kern 3.2pc$ \cmark} & \multicolumn{2}{l}{\qquad \cmark $\kern 2.1pc$ \xmark} \\ 
% % \hline 
% GQA~\cite{gqa_dataset} & \multicolumn{3}{l}{\qquad \cmark $\kern 2.3pc$ \xmark $\kern 2.7pc$ \xmark} & \multicolumn{5}{l}{\qquad \cmark $\kern 2.7pc$ \xmark $\kern 2.8pc$ \cmark $\kern 2.8pc$ \xmark $\kern 3.2pc$ \cmark} & \multicolumn{2}{l}{\qquad \cmark $\kern 2.1pc$ \xmark} \\ 
% % \hline
% \begin{tabular}[c]{@{}l@{}}Visual \\ Genome~\cite{visual_genome}\end{tabular}  & \multicolumn{3}{l}{\qquad \cmark $\kern 2.3pc$ \xmark $\kern 2.7pc$ \xmark} & \multicolumn{5}{l}{\qquad \cmark $\kern 2.7pc$ \xmark $\kern 2.8pc$ \cmark $\kern 2.8pc$ \xmark $\kern 3.2pc$ \xmark} & \multicolumn{2}{l}{\qquad \cmark $\kern 2.1pc$ \xmark} \\ 
% \hline
CLEVRER~\cite{clevrer} & \multicolumn{3}{l|}{\qquad\qquad \cmark $\kern 2.8pc$ \cmark $\kern 2.6pc$ \cmark} & \multicolumn{3}{l}{\qquad \xmark $\kern 2.8pc$ \xmark $\kern 3.2pc$ \cmark}\\ 
% \hline
Next-QA~\cite{next_qa_dataset} & \multicolumn{3}{l|}{\qquad\qquad \xmark $\kern 2.9pc$ \cmark $\kern 2.6pc$ \cmark} & \multicolumn{3}{l}{\qquad \xmark $\kern 2.8pc$ \cmark $\kern 3.2pc$ \cmark}\\ 
% \hline
ActivityNet-QA~\cite{activity_net_dataset} & \multicolumn{3}{l|}{\qquad\qquad \xmark $\kern 2.9pc$ \xmark $\kern 2.6pc$ \cmark} & \multicolumn{3}{l}{\qquad \xmark $\kern 2.8pc$ \cmark $\kern 3.2pc$ \xmark}\\ 
% \hline
STAR~\cite{star_situated_reasoning} & \multicolumn{3}{l|}{\qquad\qquad \cmark $\kern 2.7pc$ \cmark $\kern 2.6pc$ \cmark} & \multicolumn{3}{l}{\qquad \xmark $\kern 2.8pc$ \cmark $\kern 3.2pc$ \cmark}\\ 
% \hline
Causal-VidQA~\cite{causal_vid_qa} & \multicolumn{3}{l|}{\qquad\qquad \xmark $\kern 2.8pc$ \cmark $\kern 2.6pc$ \xmark} & \multicolumn{3}{l}{\qquad \xmark $\kern 2.8pc$ \cmark $\kern 3.2pc$ \cmark}\\  
% \hline
% \begin{tabular}[c]{@{}l@{}}Action \\ Genome~\cite{action_genome}\end{tabular} & \multicolumn{3}{l|}{\qquad \xmark $\kern 2.3pc$ \xmark $\kern 2.7pc$ \xmark} & \multicolumn{5}{l|}{\qquad \xmark $\kern 2.7pc$ \cmark $\kern 2.8pc$ \cmark $\kern 2.8pc$ \cmark $\kern 3.2pc$ \xmark} & \multicolumn{2}{l}{\qquad \cmark $\kern 2.1pc$ \cmark}\\ 
% \hline
% \hline
EPIC-KITCHENS~\cite{epic_kitchens} & \multicolumn{3}{l|}{\qquad\qquad \cmark $\kern 2.7pc$ \cmark $\kern 2.6pc$ \xmark} & \multicolumn{3}{l}{\qquad \cmark $\kern 2.7pc$ \cmark $\kern 3.2pc$ \cmark}\\
% \hline
Ego-4D~\cite{ego_4d} & \multicolumn{3}{l|}{\qquad\qquad \xmark $\kern 2.8pc$ \cmark $\kern 2.6pc$ \xmark} & \multicolumn{3}{l}{\qquad \cmark $\kern 2.7pc$ \cmark $\kern 3.2pc$ \cmark}\\ 
% \hline
VIOLIN~\cite{violin_dataset} & \multicolumn{3}{l|}{\qquad\qquad \xmark $\kern 2.8pc$ \cmark $\kern 2.6pc$ \xmark} & \multicolumn{3}{l}{\qquad \xmark $\kern 2.8pc$ \cmark $\kern 3.2pc$ \xmark}\\
% \hline
Cross-Task~\cite{cross_task} & \multicolumn{3}{l|}{\qquad\qquad \cmark $\kern 2.7pc$ \cmark $\kern 2.6pc$ \cmark} & \multicolumn{3}{l}{\qquad \xmark $\kern 2.8pc$ \cmark $\kern 3.2pc$ \xmark}\\
\hline
\textbf{EgoTV}\parbox[c]{1em}{
      \includegraphics[width=0.1in]{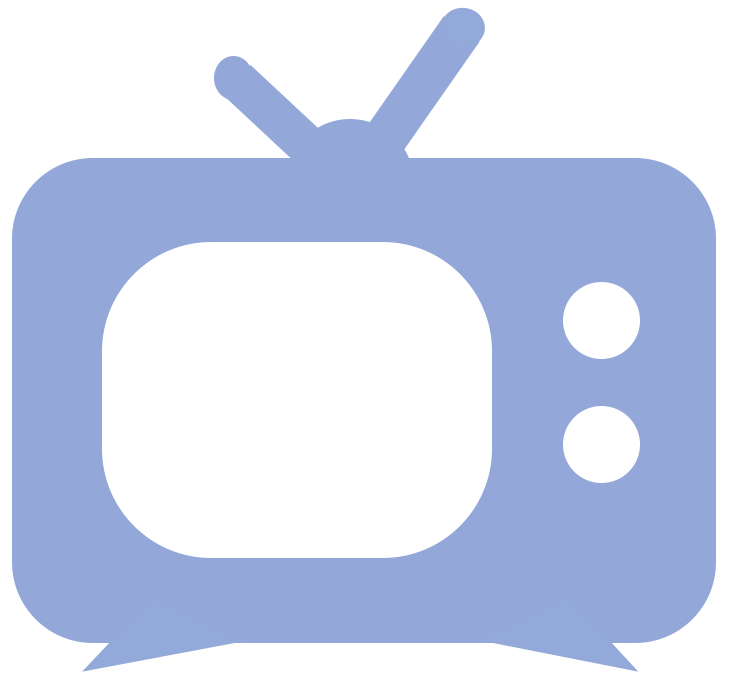}} & \multicolumn{3}{l|}{\qquad\qquad \blueD{\cmark} $\kern 2.5pc$ \blueD{\cmark} $\kern 2.3pc$ \blueD{\cmark}} & \multicolumn{3}{l}{\qquad \blueD{\cmark} $\kern 2.5pc$ \red{\xmark} $\kern 3.0pc$ \blueD{\cmark}}\\
% \vspace{2cm}
\hline
\end{tabular}
%}
% \vspace{0.5cm}
\caption{\textbf{\etv vs. existing video-language datasets.} \etv benchmark enables systematic investigation (diagnostics) on compositional, causal (e.g., effect of actions), and temporal (e.g., action ordering) reasoning in egocentric settings. Table~\ref{table:list_of_datasets_full} in Appendix provides a more comprehensive comparison.}
\label{table:list_of_datasets_abridged}
% \vspace{-10pt}
\end{table*}

% \input{figures/TV.png}
% \input{tables/Tab 1 (full)}
% spatio-temporal and compositional reasoning: CLEVRER~\cite{clevrer}, NExT-QA~\cite{next_qa_dataset}, AGQA~\cite{agqa_dataset}, ActivityNet-QA~\cite{activity_net_dataset}, STAR~\cite{star_situated_reasoning} and TVQA~\cite{tvqa_dataset}; causal reasoning: CLEVRER~\cite{clevrer}, NExT-QA~\cite{next_qa_dataset}, Social-IQ~\cite{social_iq} and Causal-VidQA~\cite{causal_vid_qa}; (ii) visual captioning on images: MSCOCO~\cite{mscoco_dataset}, Flickr30k~\cite{flickr30k} and videos: MovieQA~\cite{movie_qa}; (iii) visual entailment on images SNLI-VE~\cite{snli_ve_dataset} and videos: VIOLIN~\cite{violin_dataset}.

%% file: sections_CRV/3_Dataset.tex
\section{\etv Benchmark and Dataset}
\label{section:dataset}

We present the \emph{\textbf{Ego}centric \textbf{T}ask \textbf{V}erification} (\etv) benchmark and dataset. To enable task tracking and verification for egocentric agents, \etv contains:~1)~\emph{multi-step} tasks with \emph{ordering constraints} to capture the causal and temporal nature of everyday tasks,~2)~\emph{multimodality} -- language in addition to the egocentric video to allow language-based human-agent interaction. 

\etv also aims to enable the systematic study of generalization in task verification (see Table~\ref{table:list_of_datasets_abridged}). To this end, we create the \etv dataset using a photo-realistic simulator AI2-THOR~\cite{ai2thor} -- as a rich testbed for future research on generalizable agents for task tracking and verification. Our synthetic dataset serves as a valuable proxy of real-world performance of various task verification models while providing control over various factors affecting task reasoning. Lastly, we also create a real-world task verification dataset (\S~\ref{section:cross_task_construct}) using the CrossTask dataset~\cite{cross_task}. While this dataset is not egocentric and is limited in its ability to systematically evaluate the generalization of task reasoning models, it enables the testing of task verification models in real world.

% we provide an abridged study of \etv on real-world data  .

%%%%%%%%%%%%%%%%%%%%%%%%%%%%%%%%%%%%%%%%%%%%%%%%%%%%%%%%%%%%%%%%%
\subsection{Definitions}
\noindent \textbf{Benchmark.} The objective is to determine if a task described in natural language has been correctly executed by the agent in a given egocentric video.

\noindent \textbf{Tasks.} 
% In real-world tasks, ordering between steps might arise because of physics of the world e.g., need to pick up a knife before slicing, or semantics of the task e.g., need to slice vegetables before frying. At the same time, certain steps may be executed in any order. Consequently, 
Each task in \etv consists of multiple \emph{partially-ordered sub-tasks} or steps. A sub-task corresponds to a single object interaction via one of the six actions:~\emph{heat, clean, slice, cool, place, pick}, and is parameterized by a~\emph{target} object of interaction\footnote{Except the \emph{place} sub-task, which is additionally parameterized by a \textit{receptacle} object, we currently limit our \etv dataset to sub-tasks involving only a single target object.}. By using the ``actionable" properties of objects in AI2-THOR~\cite{ai2thor}, we ensure that the sub-tasks are parameterized with appropriate target objects in \etv, e.g., \emph{heat(book)} will never occur.

Real-world tasks consist of sub-tasks with ordering constraints, either due to physical restrictions (e.g., picking up a knife before slicing) or task semantics (e.g., slicing vegetables before frying).
We allow \etv tasks to be partially ordered, with some steps following strict ordering, e.g.~\emph{pick} sub-task happens before \emph{place} sub-task, while others are order-independent.

The ordering constraints between sub-tasks are captured in the task description using specifiers such as \emph{and}, \emph{then}, and \emph{before/after}. For simplicity, we will refer to a task using $\left< \text{\emph{sub-task}} \right> \_ \left< \text{\emph{ordering-specifier}} \right>$ notation, irrespective of the actual task description. Such tasks can then be instantiated by specifying an $(object)$ of interaction. An example task instance from \etv: ~\emph{heat\_then\_clean(apple)} is shown in Fig.~\ref{figure:dataset} with its NL description: ``apple is heated, then cleaned in a sinkbasin". The task consists of two ordered sub-tasks: heat $\rightarrow$ clean on \emph{target} object: apple. We adopt this terminology from ALFRED~\cite{ALFRED20}.

%%%%%%%%%%%%%%%%%%%%%%%%%%%%%%%%%%%%%%%%%%%%%%%%%%%%%%%%%%%%%%
\subsection{Dataset} As shown in Fig.~\ref{figure:dataset}, \etv dataset consists of (task description, video) pairs with positive or negative task verification labels. By combining the six sub-tasks~\emph{heat, clean, slice, cool, put, pick} with different ordering constraints, we create 82 tasks for \etv (see Appendix~\ref{appendix:dataset_analysis_and_statistics} for an exhaustive list). Tasks are instantiated with 130 target objects (excluding visual variations in shape, texture, and color) and 24 receptacle objects, totaling 1038 task-object combinations. These are performed in 30 different kitchen scenes. We also provide comprehensive annotations for each video, including frame-by-frame breakdowns for sub-tasks, object bounding boxes, and object state information (e.g.,~\textit{hot, cold, etc.}) to facilitate future research.

%================================================================
\subsubsection{Generation}

\noindent \textbf{Task-video Generation.} We generate the videos in our dataset by leveraging the ALFRED setup~\cite{ALFRED20}. ALFRED allows us to specify the \etv tasks using Planning Domain Definition Language (PDDL) and then to generate plans for achieving these tasks using the Metric-FF planner~\cite{metric_ff}. We execute these plans using the AI2-THOR simulator and obtain their corresponding videos. Further details on encoding tasks using PDDL and planning are in Appendix~\ref{appendix:etv_taskvideogen}.

\noindent \textbf{Task-description Generation.} We convert the plans generated for each task into positive and negative task descriptions using templates. Appendix~\ref{appendix:task_templates} provides details on the process and example templates.

\begin{figure}[t]
\centering
    \includegraphics[width=\linewidth]{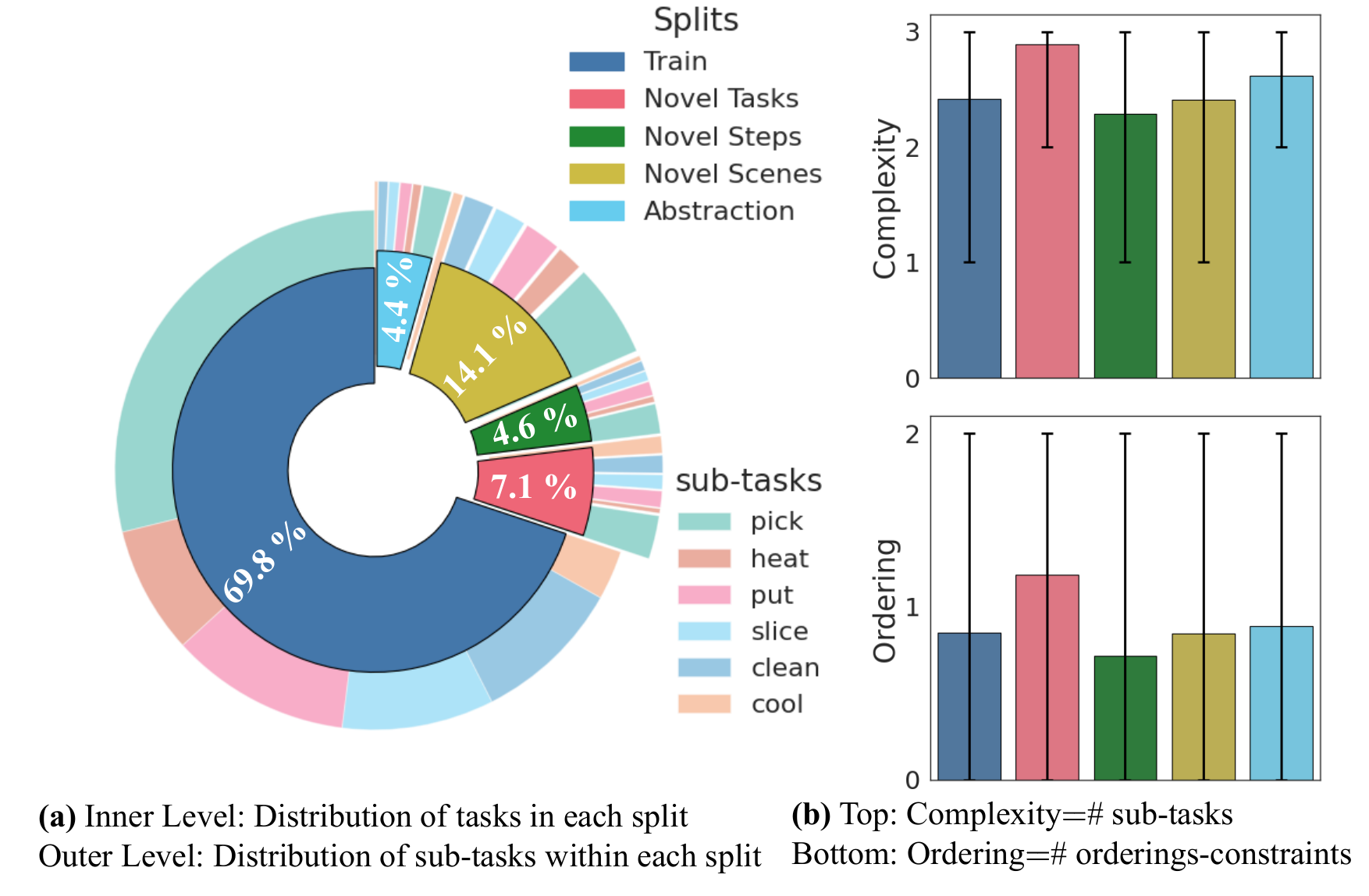}
    \caption{\textbf{\etv dataset.} Sub-tasks and tasks, including their difficulty measures (\S~\ref{section:evaluation}) are shown per split. Novel Scenes have more tasks since all the train tasks are repeated in unseen scenes. Likewise, complexity and ordering are higher in Novel Tasks due to the addition of unseen sub-tasks.}
    \label{figure:stats}
    % \vspace{-0.3cm}
\end{figure}

%=====================================================================

\subsubsection{Evaluation} 
\label{section:evaluation}

\noindent \textbf{Metrics.}
We use accuracy and F1 to measure the efficacy of models on \etv task verification benchmark. To capture the difficulty of tracking and verifying tasks, we introduce two measures:~(1)~\emph{Complexity}: measuring the number of sub-tasks in a task, which impacts the video length and requires higher action and object grounding, and~(2)~\emph{Ordering}: measuring the number of ordering constraints in a task and measures the difficulty of temporal reasoning required to track and verify tasks. We evaluate model scalability by testing on tasks with varying complexity and ordering.

\noindent \textbf{Generalization.}
\etv dataset enables systematic exploration of generalization in task tracking and verification via four test splits that focus on generalization to novel steps, tasks, visual contexts/scenes, and abstract task descriptions.

\begin{itemize}[leftmargin=*,noitemsep]
    \item \textbf{\nt}: Unseen compositions of seen sub-tasks. For e.g., if train set is \{\textit{clean(apple)},~\textit{cool(apple)}\}, then this test split would contain tasks like: \{\textit{clean\_and\_cool(apple)},~\textit{clean\_then\_cool(apple)}, \textit{ cool\_then\_clean(apple)}\}.

    \item \textbf{\nst}: Unseen compositions of sub-task actions and target objects. For e.g., if the train set is \{\textit{clean(apple)},~\textit{cool(egg)},~\textit{clean\_and\_cool(tomato)}\}, then this test split would contain tasks like: \{\textit{clean(egg)},~\textit{cool(apple)},~\textit{clean\_and\_cool(apple)}\}.

    \item \textbf{\nc}: This test split contains the same tasks as in the train set. However, the tasks are executed in unseen kitchen scenes.

    \item \textbf{Abstraction}: Abstract task descriptions, which lack the low-level details of the task. For instance, for a \textit{heat\_and\_clean(apple)} task, the full task description in the train set could be ``apple is heated in a microwave and cleaned in sink basin", while the abstract task description in this split could be ``apple is heated and cleaned".
\end{itemize}

Note that all the test splits and the train set are disjoint from each other. \nst split tests an \etv model's ability to understand generalizable object affordances and tool usage. For instance, once a model learns the \emph{slice} action on an apple, this split tests if the model can apply it to an orange. On the other hand, the \nt split tests the generalization of a model's temporal and causal reasoning capabilities on unseen compositions and orderings of known sub-tasks. Existing real-world datasets like Ego4D~\cite{ego_4d} and EPIC-KITCHENS~\cite{epic_kitchens} fail to provide such systematic control and precise diagnostics across various relevant yet independent factors affecting task reasoning.

\subsubsection{Statistics}
\etv dataset consists of 7,673 samples (train set: 5,363 and test set: 2,310). The split-wise division is \nt: $540$, \nst: $350$, \nc: $1082$, Abstraction: $338$. The total duration of the egocentric videos in the \etv dataset is 168 hours, with an average video length of 84 seconds. To ensure diversity, each task in \etv is associated with $\approx$10 different task description templates (inclusive of positive and negative scenarios). We also keep an additional template set for the abstraction split. The task descriptions consist of 9 words on average, with a total vocabulary size of 72. On average, there are 4.6 sub-tasks per task in the \etv dataset, and each sub-task spans approximately 14 frames. Additionally, there are 2.4 ways to verify a task. This requires the virtual agent to understand all possible temporal orderings between sub-tasks from the task description for successful task verification. Real-world datasets mainly focus on recognizing actions, objects, and state changes~\cite{ego_4d, epic_kitchens} without this ambiguity. Figure~\ref{figure:stats} shows a comparison of train and test splits (more analysis in Appendix~\ref{appendix:dataset_analysis_and_statistics}).

%%%%%%%%%%%%%%%%%%%%%%%%%%%%%%%%%%%%%%%%%%%%%%%%%%%%%%%%%%%%%%%%%%%%%%%%%%%%

%% file: sections_CRV/3.5_CTV.tex
% \input{tables/task_topics}

\section{CrossTask Verification (CTV) Dataset}
\label{section:cross_task_construct}

Drawing from the \etv dataset, we introduce CrossTask Verification (CTV) dataset, using videos from the CrossTask dataset~\cite{cross_task}, to evaluate task verification models on real-world videos. In CTV, we prioritize assessing real-world performance of task verification models over systematic study of their generalization capabilities, unlike \etv. Thus, CTV complements \etv dataset -- CTV and \etv together provide a solid test-bed for future research on task verification.

% Example tasks are shown in Table~\ref{table:task_topics}. Notably, some actions, like \textit{add sugar}, are common across tasks such as \textit{Make lemonade} and \textit{Make coffee}. The temporal annotation of each action is available. 
% In CrossTask Verification evaluation, given a task description and the video, the model needs to predict if the task is accomplished (entailed) in the video. 

% \input{tables/query_definition}
% Like \etv, CTV consists of paired task descriptions and videos for task verification. We construct two settings: (1) \textbf{Action sequence verification.} NL descriptions are obtained by leveraging action step annotations in CrossTask. The model's objective is to determine whether the action steps are carried out correctly and sequentially. (2) \textbf{Task~verification.} Task class labels are used as descriptions. Here, the model requires a broader understanding of the task as a whole. The dataset construction details are provided in Appendix~\ref{appendix:cross_task_construct}
%%%%%%%%%%%%%%%%%%%%%%%%%%%%%%%%%%%%%%%%%%%%%
\subsection{Dataset Generation}
\label{subsection:cross_task_construct_problem}
\noindent Like \etv, CTV consists of paired task descriptions and videos for task verification. CrossTask has 18 task classes, each with roughly 150 videos, from which we create $\approx$ 2.7K samples. We generate task descriptions by concatenating action step annotations in CrossTask. The model's objective is to determine whether the action steps (sub-tasks) and their sequence in the video align with the description. 
%(2) \textbf{Task~verification.} Task class labels are used as descriptions. Here, the model requires a broader understanding of the task as a whole. 
See Appendix~\ref{appendix:cross_task_construct} for dataset construction details.
% Additional dataset construction details are provided in Appendix~\ref{appendix:cross_task_construct}.
\input{figures/cross_task_verification}

\subsection{Evaluation}
\noindent \textbf{Metrics.}
Following \etv, we use accuracy and F1 to measure the efficacy of the models on the CTV dataset. %We also evaluate the model's scalability on tasks with varying complexity and ordering.

\noindent \textbf{Generalization.}
%For \textit{action sequence verification}, 
We construct a test set using videos with seen action steps but in previously unseen compositions. To ensure novel compositions, we train on videos with up to 3 action steps and test on those with 4, as illustrated in Figure \ref{figure:cross_task_verification}. While this mirrors the Novel Task split in \etv, the CTV test set also contains unseen visual contexts (videos) -- a result of limited control during dataset creation. %Note that every sub-task in the test set is also present in the training set to ensure that all steps are observed (unseen compositions of seen sub-tasks). 
%For \textit{task verification}, we follow the same training and testing videos in the action sequence verification. The difference is that the task description was substituted from the action sequence to the task class name.
%For \textit{task verification}, similar to the previous setting, we also ensure the task description (task class) where seen and the sub-task composition in the video is unseen. 

%% file: figures/cross_task_verification.tex
% \begin{figure}[t]
% \centering
%     \includegraphics[width=\linewidth]{plots/cross_task_verification.png}
%     \caption{CrossTask Verification}
%     \label{figure:cross_task_verification}
% % \vspace{-0.3cm}
% \end{figure}

\begin{figure}[t]
\centering
    \includegraphics[width=0.9\linewidth]{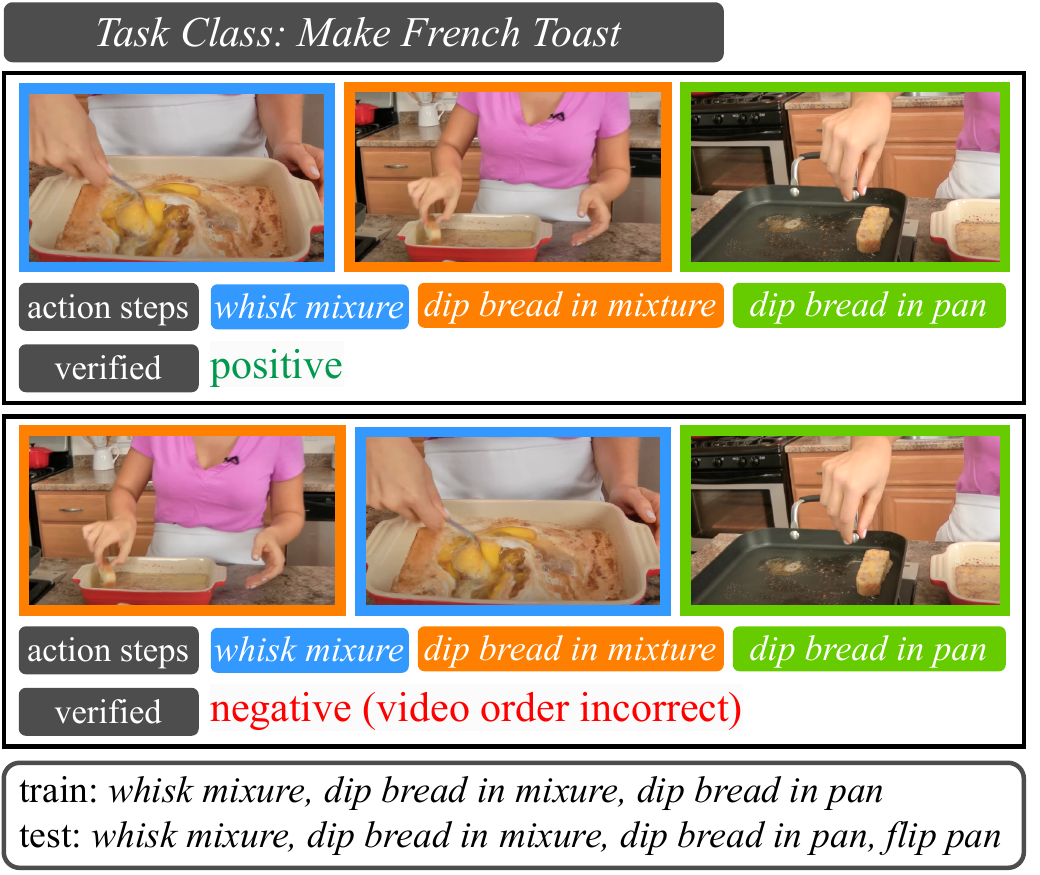}
    \caption{\textbf{CrossTask Verification (CTV) dataset.}}
    \label{figure:cross_task_verification}
     % \vspace{-0.3cm}
\end{figure}

%% file: sections_CRV/4_NSG_for_EgoTV.tex
\input{tables/query_definition}
%%%%%%%%%%%%%%%%%%%%%%%%%%%%%%%%%%%%%%%%%%%%%%

\section{Neuro-Symbolic Grounding (NSG)}
\label{section:proposed_framework}

\etv requires visual grounding of task-relevant entities such as actions, state changes, etc. extracted from NL task descriptions for verifying tasks in videos. To enable grounding that generalizes to novel compositions of tasks and actions, we propose the Neuro-symbolic Grounding (NSG) approach. NSG consists of three modules:~a)~semantic parser, which converts task-relevant states from NL task descriptions into symbolic graphs,~b)~query encoders, which generate the probability of a node in the symbolic graph being grounded in a video segment, and~c)~video aligner, which uses the query encoders to align these symbolic graphs with videos. NSG thus uses intermediate symbolic representations between NL task descriptions and corresponding videos to achieve compositional generalization.

%%%%%%%%%%%%%%%%%%%%%%%%%%%%%%%%%%%%%%%%%%%%%%%%%
\subsection{Queries for Symbolic Operations}
\label{subsection:queries}

To encode tasks, NSG captures task-relevant visual and relational information in a structured manner via symbolic operators called \emph{queries}.
For instance, the task description \emph{heat an apple} can be symbolically captured by the query: \code{StateQuery(apple, hot)}.
Similarly, the task description \emph{place steak on grill} can be captured by \code{RelationQuery(steak, grill, on)}, which represents the relation (\code{on}) between objects \code{steak} and \code{grill}.
Queries are characterized by \textit{types} and \textit{arguments} and are stored in a text format. Table~\ref{table:QTypes} shows the various query types and their arguments. Different query types capture different aspects, e.g., attributes, relations, etc., thereby enabling a rich symbolic representation of everyday tasks.

\subsection{Semantic Parser for Task Descriptions}
\label{subsection:plan_parsing_from_instructions}

The symbolic operators, i.e., queries, allow the semantic parser to represent a task's partial-ordered steps using a symbolic graph. Specifically, the parser translates a NL task description into a graph $G(V, E)$, where a vertex $n_i \in V$ represents a query and an edge $e_{ij}: n_i \rightarrow n_j \in E$ is an ordering constraint indicating that $n_i$ must precede $n_j$~(Figure~\ref{figure:model-layout}a). We experiment with two different methods to parse language descriptions of tasks to graphs -- (i) finetuning language models and (ii) few-shot prompting of language models. For details, refer to Appendix~\ref{section:appendix_semantic_parsing}. We perform a topological sort with the graph $G$ and generate all the possible sequences of queries consistent with the sort. For example, the topological sorting of the graph in Figure~\ref{figure:model-layout}(a) yields two ordered sequences: $(n_0, n_1, n_2, n_3)$, $(n_0, n_2, n_1, n_3)$. Note that this does not include all physically possible ways to complete a task, but a super-set of all possible sequences of task-relevant queries, including some infeasible sequences\footnote{For instance, in Figure~\ref{figure:model-layout}a, $n_1$ and $n_2$ are at the same topological level, but the sub-task in query $n_1$ could invalidate pre-conditions for $n_2$. Hence, a physically plausible task requires $n_2$ followed by $n_1$ and not vice versa. Note that \etv does not have physically implausible tasks.}. However, this super-set is useful because a task can be verified as accomplished if any sequence in this set can be ascertained to occur in the video. 

Notably, all \etv tasks map to acyclic graphs through temporal disambiguation. While this can support tasks with repeated actions, such as: (Task) \textit{pick two apples}; (Graph) pick(apple) $\rightarrow$ pick(apple); tasks that require (recursively) repeating action sequences until a desired state is reached, might result in cyclic graphs.  Examples include unstacking an arbitrary number of dishes or searching for an ingredient. While currently absent in \etv, extending to such tasks would be a valuable future direction.

\subsection{Query Encoders for Grounding}
\label{subsection:query_encoders}

Query Encoders are neural network modules that evaluate whether a query is satisfied in an input video. Specifically, a query encoder $f^{\theta_{\tau}}$ for a query $n$ of type $\tau$ (e.g., \code{StateQuery}, \code{RelationQuery} etc.), accepts NL arguments ($a$) corresponding to objects and relations in $n$ and a video ($v$) to generate the probability $\mathds{P}=f^{\theta_{\tau}}(a, v)$ of the desired query being true in the video. Learnable parameters corresponding to different query type encoders in an NSG model are jointly represented as $\theta = \bigcup_{\tau}\theta_{\tau}$.

Both the text arguments $a$ of the query and the frames of the input video $v$ are encoded using a pre-trained CLIP encoder~\cite{clip}. The token-level and frame-level representations from CLIP are separately aggregated using two LSTMs~\cite{lstm} to obtain aggregated features for $a$ and $v$, respectively. These features are then fused and passed through the neural network $f^{\theta_{\tau}}$ to obtain the probability $\mathds{P}$ of the query being true in the video (see Figure~\ref{figure:model-layout}a).
%%%%%%%%%%%%%%%%%%%%%%%%%%%%%%%%%%%%%%%%%%%%%%%%%%%%%%%%
\input{figures/model-layout}
%%%%%%%%%%%%%%%%%%%%%%%%%%%%%%%%%%%%%%%%%%%%%%%%%
\subsection{Video Aligner for Task Verification}
\label{subsection:plan_verification}
% say we formulate as a query alignment problem
This module of NSG must align the graph representation $G$ of the task (generated by the semantic parser) with the video. To that end, it first segments the video, then jointly learns~a)~the query encoders, which detect the queries in the video segments and~b)~the alignment between video segments and the query sequences obtained from the topological sort on $G$. Such joint learning is required since the temporal locations of the queries in the video are unknown a priori requiring simultaneous detection and alignment. If the video is a positive match for the task encoded in $G$, at least one of the query sequences from $G$ must temporally align perfectly with the video segments for successful task verification. Conversely, for negative matches, no query sequence from $G$ would \emph{completely} align with the video segments. Going forward, we use $\langle\rangle$ and $()$ to denote ordered pairs and sequences, respectively.

\noindent \textbf{Video Segmentation:} The video is segmented into non-overlapping segments\footnote{Since pretrained, off-the-shelf video segmentation models are limited to predefined action classes~\cite{escorcia2016daps} or reliant on background frame change detection~\cite{yang2022temporal} and require downstream finetuning~\cite{gao2020accurate}, we leave their integration in NSG as future work.} with a moving window of arbitrary, but fixed size $k$\footnote{If required, the last segment is zero-padded to $k$ frames.} 

\noindent \textbf{Joint Optimization:} The objective of the optimization is to jointly learn the \emph{alignment} $\mathrm{Z}$ between queries and video segments along with the \emph{query encoders} $f^{\theta}$.
% that detect queries in the segments and output a task verification probability for \etv tasks.
% , using only the ground truth label $y$ as supervision.
Given:~a)~the temporal sequence of $S$ segments $(s_t)_{t=0}^{S-1}$ with each $s_t$ spanning $k$ image frames; and~b)~a sequence of $N$ queries $(n_j)_{j=0}^{N-1}$  from the topological sort on $G$, the alignment $\mathrm{Z}$ is defined as a matrix $\mathrm{Z} \in \{0, 1\}^{N \times S}$, where $Z_{jt} = 1$ implies that the $j^{th}$ query $n_j$ is aligned the video segment $s_t$. An example alignment with $N=2$ and $S=3$ is given by the matrix $\mathrm{Z} = \left[ \begin{array}{ccc} 1 & 0 & 0 \\ 0 & 0 & 1 \end{array} \right]$, where the rows are ordered queries $(n_0, n_1)$, the columns are temporal segments $(s_0, s_1, s_2)$, and $\langle n_0, s_0 \rangle$, $\langle n_1, s_2 \rangle$ are the aligned pairs. Assuming segmentation guarantees sufficient segments for query alignment: $S \geq N$. Using $\mathrm{Z}$ and $f^{\theta}$, the task verification probability $p^{\theta}$ can be defined as:
% \vspace{-0.3cm}
\begin{align}
     p^{\theta} = \sigma \bigg(\max_{\mathrm{Z} \in {\{0,1\}}^{N \times S}} \frac{1}{N}\sum_{j,t} \log f^{\theta}(a_j, s_t)Z_{jt}\bigg) \label{eq:p_theta} 
     %\vspace{-0.2cm}
\end{align}

Here $\sigma$ is the sigmoid function, $f^{\theta}(a_j, s_t)$ denotes the probability of querying segment $s_t$ using query $n_j$ with arguments $a_j$ (\S~\ref{subsection:query_encoders}), and $\max$ operator is over the best alignment $\mathrm{Z}$ between $N$ queries and $S$ segments. We use the ground-truth task verification label $y$ to compute $\mathrm{Z}$ and $f^{\theta}$ by minimizing the following loss:
\vspace{-0.2cm}
\begin{align}
     \min_{\theta} & \frac{1}{|\mathcal{D}|} \sum \mathcal{L}_{\text{BCE}}(p^{\theta}, y), \label{eq:bce}
    %\vspace{-0.2cm}
\end{align}

here $|\mathcal{D}|$ is the \etv dataset size and $\mathcal{L}_{\text{BCE}} (\cdot)$ is the binary cross entropy loss computed over $|\mathcal{D}|$ input,~output pairs.~Given the minimax nature of Eq.~\ref{eq:bce}, we use a 2-step iterative optimization process:~(i)~find the best alignment $\mathrm{Z}$ between queries and segments with fixed query encoder parameters $\theta$ (optimize Eq.~\ref{eq:p_theta} with fixed $f^{\theta}$);~(ii)~optimize $\theta$ using Eq.~\ref{eq:bce}, given $\mathrm{Z}$.

%%%%%%%%%%%%%%%%%%%%%%%%%%%%%%%%%%%%%%%%%%%%%%%%%%%%%%%%%

\noindent \textbf{Dynamic Programming (DP)-based Alignment:} Finding the best $\mathrm{Z}$ in Eq.~\ref{eq:p_theta} given $\theta$ requires iterating over combinations of $N$ queries and $S$ segments while respecting certain constraints. The constraints, visualized in Fig.~\ref{figure:model-layout}b, ensure that~a)~no two queries are aligned to the same segment\footnote{This ensures that the order of queries can be verified, which cannot be done when queries belong to the same segment.} (Eq.~\ref{Z-a}),~b)~all queries are accounted for in $S$ (Eq.~\ref{Z-b}), and~c)~the temporal orderings between queries in the query sequences are respected (Eq.~\ref{Z-c}). Specifically, if query $n_u$ precedes $n_v$ ($n_u \rightarrow n_v$), and query $n_v$ is paired with segment $s_{\bar{t}}$ (i.e. $Z_{v\bar{t}}=1$), then query $n_u$ cannot be paired with any segment that lies after $s_{\bar{t}}$ (i.e. $Z_{ut} \neq 1 \; \forall \; t \geq \bar{t}$). The resulting optimization problem for $\mathrm{Z}$, given $\theta$ is:
% \vspace{-0.2cm}
\begin{subequations}\label{Z-main}
\begin{align}
& \max_{\mathrm{Z} \in {\{0,1\}}^{N \times S}} \sum_{j,t} \log f^{\theta}(a_j, s_t)Z_{jt} \tag{\ref{Z-main}}, \quad \text{s.t.}\\
& \sum_{j=0}^{N-1} Z_{jt} \in \{0,1\}, \quad \forall \; 0 \leq t \leq S-1 \label{Z-a}\\
&\sum_{t=0}^{S-1} Z_{jt} = 1, \quad \forall \; 0 \leq j \leq N-1 \label{Z-b}\\
% & n_u \rightarrow n_v, \; Z_{v\bar{t}}=1 \Longrightarrow Z_{ut} \neq 1, \quad \forall \; t \geq \text{argmax}_{\bar{t}} Z_{v\bar{t}} \label{Z-c}
& n_u \rightarrow n_v, \; Z_{v\bar{t}}=1 \Longrightarrow Z_{ut} \neq 1, \quad \forall \; t \geq \bar{t} \label{Z-c}
\end{align}
\end{subequations}

Intuitively, the solution to Eq.~\ref{Z-main} gives us the best alignment score (note, the overlap with Eq.~\ref{eq:p_theta}). The iterations over $N$ queries and $S$ segments for solving Eq.~\ref{Z-main} are underpinned by an overlapping and optimal substructure. For instance, to optimally align queries $(n_j)_{j=0}^{N-1}$ and segments $(s_t)_{t=0}^{S-1}$, one could:~a)~pair $\langle n_0, s_0 \rangle$ and optimally align the remaining queries and segments $(n_j)_{j=1}^{N-1}, (s_t)_{t=1}^{S-1}$; or (2) skip $s_0$ and still optimally align \emph{all} queries, now with the remaining segments $(n_j)_{j=0}^{N-1}, (s_t)_{t=1}^{S-1}$ (see Fig.~\ref{figure:model-layout}b(iv)). This recursive substructure leads to a DP solution for Eq.~\ref{Z-main}. 

Let, $F^{\ast}((n_j)_{\bar{j}}^{N-1}, (s_t)_{\bar{t}}^{S-1})$ denote the best alignment score for queries $(n_j)_{\bar{j}}^{N-1}$ and segments $(s_t)_{\bar{t}}^{S-1}$ from Eq.~\ref{Z-main}. Based on the aforementioned reasoning, $F^{\ast}((n_j)_{\bar{j}}^{N-1}, (s_t)_{\bar{t}}^{S-1})$ can be recursively written as: 
% \vspace{-0.2cm}
\begin{multline}
    F^{\ast}((n_j)_{\bar{j}}^{N-1}, (s_t)_{\bar{t}}^{S-1}) = \text{max} \big( \log f^{\theta}(a_{\bar{j}}, s_{\bar{t}}) \\ + F^{\ast}((n_j)_{\bar{j}+1}^{N-1}, (s_t)_{\bar{t}+1}^{S-1}), F^{\ast}((n_j)_{\bar{j}}^{N-1}, (s_t)_{\bar{t}+1}^{S-1}) \big) \label{eq:dp}
\end{multline}

The base cases for the DP are: (i) $\mathrm{Z}=\mathds{I} \; \text{if} \; N=S$; (ii) $Z_{jt} = 1 \; \forall \; t \; \text{if} \; j=N-1$. It is worth noting that the DP subproblems, together with the base cases, satisfy the constraints in Eq.~\ref{Z-a}~\ref{Z-b}~\ref{Z-c}. Since the video may match any of the sequence in the super-set of query sequences (from the topological sort on $G$), we repeat this process of computing $F^{\ast}$ for each sequence and select the maximum value.

%%%%%%%%%%%%%%%%%%%%%%%%%%%%%%%%%%%%%%%%%%%%%%%%%%%%%%%%%5
\noindent \textbf{Optimizing Query Encoder Parameters $\theta$:} After obtaining the best alignment $\mathrm{Z}$ using DP, we substitute the corresponding value of $F^{\ast}((n_j)_{j=0}^{N-1}, (s_t)_{t=0}^{S-1})$ in Eq.~\ref{eq:p_theta} and subsequently Eq.~\ref{eq:bce}. In Eq.~\ref{eq:bce}, we use single mini-batch of training examples and take one gradient-update step of the Adam optimizer for the query encoder parameters $\theta$.

%%%%%%%%%%%%%%%%%%%%%%%%%%%%%%%%%%%%%%%%%%%%%%%%%%%%%%%%%%%%%%%%%%%%%%%%%%%%%%%%%

\input{tables/baseline_results.tex}

%% file: tables/query_definition.tex
\begin{table*}[t]
\centering
\footnotesize{
\begin{tabular}{l|l|l}
    \thead{Query Type} & \thead{Signature} & \thead{Semantics} \\
    \hline
     \code{StateQuery} & \makecell[l]{(Object, State), Video $\mapsto \mathds{P}$} & \makecell[l]{Queries the state (\code{hot}, \code{cold}, \code{clean}, \code{ripe}) of object in a video\\and returns the probability of the object state being detected.\\ Example instructions: \emph{heat an apple, clean a spoon}.} \\
     \hline
     \code{RelationQuery} & \makecell[l]{(Object, Object/Receptacle, Relation), Video $\mapsto \mathds{P}$} & \makecell[l]{Queries the relation between two objects or an object and a\\ receptacle in a video and returns the probability of the relation \\
     being detected. Example instructions: \textit{put apple in basket}, \\ \textit{place spoon to the left of plate}.} \\
     \hline
     \code{ActionQuery} & \makecell[l]{(Subtask, \textsuperscript{$\ast$}Objects, \textsuperscript{$\ast$}Relation), Video $\mapsto \mathds{P}$} & \makecell[l]{Queries for a sub-task with one or more arguments ($\ast$) in a video \\and returns the probability of the sub-task being executed. \\ Example instructions: \emph{whisk mixture, pour lemonade into glass}.} \\
     % \hline
     % \code{AttributeQuery} & \makecell[l]{(Object, Attribute), Video $\mapsto$ Video} & \makecell[l]{Queries for object attributes like shape, color, and size in a video \\and return the video annotated with objects with same attribute. \\This could be further decomposed to \code{\{Color/Shape/Size\}Query}. \\Example instructions: \emph{grab a round ladle}, \emph{slice a green apple}}\\
     % \hline
     % \code{PropertyQuery} & (Property), Video $\mapsto$ Video & \makecell[l]{Queries for properties/affordances of objects in a video and \\ returns the video annotated with objects with same property. \\ Example instructions: \emph{screw wheel, lift the car}}\\ 
     % \hline
     % \code{CountQuery} & (Object, Count), Video $\mapsto$ & \makecell[l]{Queries for multi-item repetitive sub-tasks.\\ Example instructions: \emph{pick two apples, slice three tomatoes}\\ \red{RH: some confusion regarding sub-task repetition, how about} \\
     % \red{``heat 2 tomatoes"}}\\
     \hline
\end{tabular}
\vspace{2mm}
}
\caption{\textbf{NSG's query types for task verification in \etv and CTV.} The query types \code{StateQuery} and \code{RelationQuery} are used in \etv, whereas \code{ActionQuery} is used in CrossTask. Each query type $\tau$ is modeled using a neural network $f^{\theta_\tau}$ accepts unique arguments ($a$) and video frames ($v$) as input and generates an output probability $\mathds{P}=f^{\theta_{\tau}}(a, v)$ of the \textit{query} being true in the video $v$.}
\label{table:QTypes}
% \vspace{-2mm}
\end{table*} 

%% file: figures/model-layout.tex
% \begin{figure*}[t]
% 	\centering
% 	\begin{subfigure}[t]{0.60\textwidth}
% 		\centering
% 		\includegraphics[width=\linewidth]{plots/model.pdf}
% 		\caption{\red{need caption; dotted lines to denote frozen model parameters.}}
%         \label{figure:model-layout}
% 	\end{subfigure}%
% 	~ 
% 	\begin{subfigure}[t]{0.36\textwidth}
% 		\centering
% 		\includegraphics[width=\linewidth]{plots/constraints.pdf}
% 		\caption{Query alignment constraints [3a]: eq~\ref{Z-a} At most one query per segment; [3b]: eq~\ref{Z-b} All queries must be aligned; [3c]: eq~\ref{Z-c} Ordering constraints between queries obtained from topological sorting. \red{explain the last figure;}}
%         \label{figure:constraints}
% 	\end{subfigure}%
% 	% \caption{}
% 	% \label{figure:model-layout}
% \end{figure*}

\begin{figure*}
\centering
    \includegraphics[width=\linewidth]{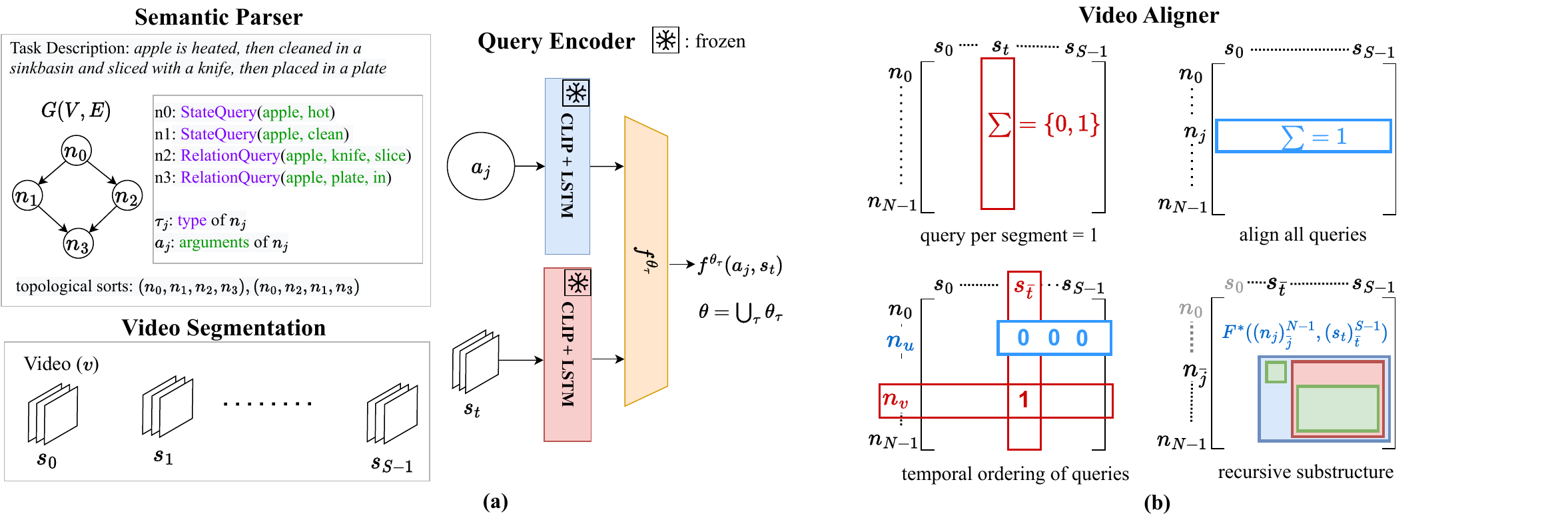}
    \caption{\textbf{NSG model}~(a)~semantic parser converts NL descriptions into a graph $G$ of symbolic queries; query encoders $f^{\theta_{\tau}}$ detect queries in individual video segments $s_t$; and a video aligner aligns $G$ with video segments by computing alignment matrix $\mathrm Z$ via a constrained optimization problem (Eq.~\ref{Z-main}).~(b)~The constraints (Eqs.~\ref{Z-a}~\ref{Z-b}~\ref{Z-c}) and the recursive structure (Eq.~\ref{eq:dp}) enabling use of DP to solve for $\mathrm Z$. Here, the blue box denotes $F^{\ast}((n_j)_{\bar{j}}^{N-1}, (s_t)_{\bar{t}}^{S-1})$, the green boxes denote $ \log f^{\theta}(a_{\bar{j}}, s_{\bar{t}}) + F^{\ast}((n_j)_{\bar{j}+1}^{N-1}, (s_t)_{\bar{t}+1}^{S-1})$, and the red box denotes $F^{\ast}((n_j)_{\bar{j}}^{N-1}, (s_t)_{\bar{t}+1}^{S-1})$.}
    
    % (a) Semantic Parser parses the language instruction into a graph $G(V, E)$ where vertices represent queries and edges represent ordering constraints. Each query has a \emph{type} and a set of arguments ($a$) and is modeled using a Query Encoder $f^{\theta_{\tau}}$ corresponding to the type. The Query Encoder processes the arguments of the query $a_j$ and a video segment $s_t$ to find the probability of the arguments being true in the video segment $f^{\theta_{\tau}}(a_j, s_t)$. (b) Constraints for DP-based alignment (i) no two queries are aligned to the same segment (Eq.~\ref{Z-a}); (ii) all queries are accounted for in $S$ (Eq.~\ref{Z-b}); (iii) temporal ordering constraints between queries are respected (Eq.~\ref{Z-c}); (iv) visualization of Eq.~\ref{eq:dp}: Here, the blue box denotes $F^{\ast}((n_j)_j^{N-1}, (s_t)_t^{S-1})$, the green boxes denote $ \log f^{\theta}(a_j, s_t) + F^{\ast}((n_j)_{j+1}^{N-1}, (s_t)_{t+1}^{S-1})$, and the red box denotes $F^{\ast}((n_j)_j^{N-1}, (s_t)_{t+1}^{S-1})$. The DP computes the best alignment of the queries and segments in the blue box by computing the $\max$ over the optimal subproblems in the green and red boxes.}
    \label{figure:model-layout}
    % \vspace{-0.3cm}
\end{figure*}

%% file: tables/baseline_results.tex
\begin{table*}[t]
% \small
\centering
\begin{adjustbox}{max width=.95\textwidth}
%\footnotesize{
\begin{tabular}{lccccccccc}
\hline
\multirow{2}{*}{\textbf{Model}}  & \textbf{Visual}  & \textbf{Text} & \textbf{MM} & \textbf{\begin{tabular}[c]{@{}c@{}} Novel  \end{tabular}} & \textbf{\begin{tabular}[c]{@{}c@{}}Novel \end{tabular}} & \textbf{\begin{tabular}[c]{@{}c@{}}Novel\end{tabular}} & \multirow{2}{*}{\textbf{Abstraction}}  &  \multirow{2}{*}{\textbf{Average}} \\ 
  & \textbf{feature} & \textbf{feature}  & \textbf{Fusion} & \textbf{Tasks} & \textbf{Steps} & \textbf{Scenes} & & \\
\hline
Text2text~\cite{roberta} &  & RoBERTa &  & 64.9 & 65.8 & 66.5 & 64.7 & 65.5 \\
\hline
\multicolumn{1}{l}{CLIP Hitchhiker \cite{clip_hitchiker}} & CLIP(I) & CLIP &  & 43.9  & 66.5  & 72.2  & 13.6 & 49.1\\ 
\multicolumn{1}{l}{CLIP4Clip mean~\cite{luo2022clip4clip}}& CLIP(I) & CLIP  &  & 49.3  & 70.9 &  74.9 & 16.1 & 52.8\\
\multicolumn{1}{l}{CLIP4Clip seqLSTM~\cite{luo2022clip4clip}} & CLIP(I) & CLIP  &  &  \underline{56.2} & 73.2  & 74.6  &  17.5 & 55.4\\ 
\multicolumn{1}{l}{CoCa \cite{coca}} & Tx(I) & Tx  &  Y &  51.5 & 71.6  &  71.9 & 43.5 & 59.6 \\ 
\multicolumn{1}{l}{VIOLIN-ResNet~\cite{violin_dataset}} & ResNet(I) & BERT  & Y & 47.4 & \textbf{80.4} & \textbf{85.6} & 42.5 & 64.0\\
\hline
\rowcolor{lightgray}
\multicolumn{1}{l}{MIL-NCE~\cite{miech2020end}} & S3D(V) & Word2vec &  & 30.5 & 69.6 & 73.5 & 24.3 & 49.5\\
\rowcolor{lightgray}
\multicolumn{1}{l}{VideoCLIP~\cite{videoclip}} & Tx(V) & Tx  & Y & 29.3  & 67.6  & 77.9  & 25.6  & 50.1 \\ 
\rowcolor{lightgray}
\multicolumn{1}{l}{VIOLIN-I3D~\cite{violin_dataset}} & I3D(V) & BERT & Y & 45.6 & \underline{79.7} & 83.9 & \underline{47.6} & 64.2 \\
\hline
\multicolumn{1}{l}{\textbf{NSG (ours)}}& CLIP(I) & CLIP & Y & \textbf{90.0} & 64.7 & \underline{84.9} & \textbf{80.4} & \textbf{80.0}\\ 
\multicolumn{1}{l}{Oracle Model} &CLIP(I) & CLIP   & Y &  95.0 & 96.7 & 97.6 & 97.2 & 96.6 \\ 
\hline
\end{tabular}%}
\end{adjustbox}
% \vspace{.2cm}
\caption{\textbf{Comparison of baselines with NSG on different data splits using F1-score.} MM fusion indicates multimodal fusion of vision and text features. %Average represent the average score among the four evaluation scenarios. 
Tx indicates the Transformer as feature extractor with image (I) and video (V) backbones. Video backbone models are highlighted in gray.  \underline{Underline} indicates second-best performance. %\red{RH: to add corresponding Accuracy results in the appendix} \rd{Add gray bg color to one of the blocks to differentiate image vs. video backbones? }
%\bc{We might replace the "Tx" to the actual name, e.g., CLIP, ViT.}
} %Con indicates contrastive loss, Gen indicates generation loss, and Class indicates classification loss.} 
%\vspace{-10pt}
\label{table:baseline_results}
\end{table*}

%% file: sections_CRV/5_Experiments.tex
\section{Experiments} 
\label{subsection:baseline_evaluation}
We compare various state-of-the-art (SOTA) VLMs with NSG on the \etv benchmark (see Appendix~\ref{appendix:nsg_training} for NSG's experimental training details).

\subsection{SOTA VLM Baselines}
We investigate 6 VLMs developed for video-language tasks requiring similar reasoning as \etv. Summarized in Table~\ref{table:baseline_results}, \textbf{CLIP4Clip}~\cite{luo2022clip4clip}, \textbf{CLIP Hitchhiker}~\cite{clip_hitchiker}, \textbf{CoCa}~\cite{coca} use image backbones followed by temporal aggregation, while \textbf{VideoCLIP}~\cite{videoclip}, \textbf{MIL-NCE}~\cite{miech2020end}, and \textbf{VIOLIN}~\cite{violin_dataset} use video backbones. With the exception of CoCa, which is trained with contrastive and captioning loss, all other models are trained using contrastive loss~\cite{miech2020end}. Lastly, VideoCLIP and VIOLIN use an explicit fusion of text-vision features. For each model, we freeze all pretrained feature extractors and finetune a fully-connected probe layer, along with the temporal aggregation layers where appropriate (CLIP4Clip-LSTM, VIOLIN), using \etv's train split. 

Finally, to establish upper bounds on \etv, we instantiate: (1) a \textbf{Text2text model}, which constructs video captions using ground-truth labels for objects and actions, encodes the captions and task descriptions using (pretrained) RoBERTa model~\cite{roberta} and measures alignment using the cosine similarity score (see Appendix~\ref{appendix:upper_bound_models}), and (2) an \textbf{Oracle model}, which is trained with full supervision on sub-tasks labels and locations in addition to task verification labels.

%%%%%%%%%%%%%%%%%%%%%%%%%%%%%%%%%%%%%%%%%%%%%%%%%%%%%%%%%%%%%%%%%%%%%%%
\subsection{Results}
 In Table~\ref{table:baseline_results}, we show the performance of NSG vs. SOTA VLMs per split of \etv.~(1)~\textbf{Novel Tasks}: NSG significantly outperforms other baselines due to its ability to decompose and detect sub-tasks while using DP alignment to handle temporal constraints among them. In contrast, other baselines rely on detecting the entire task under temporal constraints, which is more challenging. Further, image-based baselines outperform video-based baselines due to their ability to capture a greater degree of compositional detail through frame-level representations.~(2) \textbf{Novel Steps}: NSG's poor performance in this split could be attributed to its low precision in the \emph{slice} sub-task (which is dominant in this split), as shown in Figure~\ref{figure:complexity-confusion_mat} [Right]. We hypothesize that since NSG only uses the aligned segments while discarding the rest, learning to utilize context from neighboring segments to capture \emph{slice} (like picking up a knife) could be a promising future direction. (3) \textbf{Novel Scenes}: Here, NSG is comparable to the best baseline VIOLIN-ResNet. Since the tasks are identical to the train split, the success of a model is contingent on the vision encoder's ability to accurately detect the same sub-tasks in unseen scenes. Consequently, models with an additional temporal aggregation layer (VIOLIN) finetuned on \etv, tend to outperform image-based models that do not have temporal aggregation (CLIP Hitchhiker) and models with frozen video features (MIL-NCE, VideoCLIP). (4) \textbf{Abstraction}: NSG significantly outperforms the baselines, primarily due to its semantic parser, which captures the underlying structure of the description and encodes the relevant concepts, such as objects and sub-tasks, to generate an (abstract) symbolic output.

%%%%%%%%%%%%%%%%%%%%%%%%%%%%%%%%%%%%%%%%%%%%%%%%%%%%%%%%%

\subsection{Analysis of NSG}
\label{subsec:analysis}

\noindent \textbf{NSG learns to localize task-relevant entities without explicit supervision.} Figure~\ref{figure:complexity-confusion_mat} shows the confusion matrix of \code{StateQuery} \& \code{RelationQuery} outputs, which capture sub-tasks, with their ground truths. The high recall demonstrates NSG's ability to localize task-relevant entities, despite being trained using only task verification labels.

\noindent \textbf{Effect of query types on NSG.} While query types with multiple entity arguments might appear capable of modeling complex dependencies amongst entities and having more expressive power, encoding multiple entities jointly using a single encoder makes the grounding problem more challenging. Hence, in practice, we found that using a combination of \code{StateQuery} \& \code{RelationQuery} types as opposed to \code{ActionQuery} (which encodes multiple entities using a single encoder) enabled better grounding and led to better performance in terms of F1-score (Table~\ref{table:query_comparison}).

\input{figures/complexity-confusion_mat}

\noindent \textbf{NSG shows consistent performance with increasing task difficulty.}
In Figure~\ref{figure:complexity-confusion_mat}, NSG's performance is minimally affected by increase in task difficulty characterized by number of sub-tasks (complexity) and ordering constraints (\S~\ref{section:evaluation}) unlike the best-performing baseline (VIOLIN-ResNet). 

\noindent \textbf{NSG is robust to segmentation window size} The effect of $k$ on NSG is minimal (Appendix~\ref{appendix:analysis}).

\noindent \textbf{NSG also enables task verification on real-world data.} NSG outperforms all competitive baselines on CTV significantly with F1-score (NSG: $\mathbf{76.3}$, CoCa: 70.9, VideoCLIP: 49.7, VIOLIN 34.7), demonstrating its causal and compositional reasoning capabilities in real-world applications (see Appendix~\ref{appendix:NSG_crosstask} for details).

\begin{table}[t]
\small
\centering
\begin{tabular}{lcccc}
\hline
\multirow{2}{*}{NSG} & \begin{tabular}[c]{@{}c@{}} Novel  \end{tabular} & \begin{tabular}[c]{@{}c@{}}Novel \end{tabular} & \begin{tabular}[c]{@{}c@{}}Novel\end{tabular} & \multirow{2}{*}{Abstract.} \\ & Tasks & Steps & Scenes & \\
\hline
\code{Action} & 78.2 & 45.6 & 70.6 & 75.5\\
\code{State+Relation} & 90.0 & 64.7 & 84.9 & 80.4\\
\hline
\end{tabular}
\vspace{2pt}
\caption{\code{(State}~+~\code{Relation)Query}~vs.~\code{ActionQuery}}
\label{table:query_comparison}
% \vspace{-5pt}
\end{table}  

\noindent \textbf{Limitations of NSG.} (1) It does not consider multiple simultaneous actions like ``picking an apple while closing the refrigerator door", (2) The assumption of equal-length video segments may be unsuitable for sub-tasks with a highly variable duration. We defer exploration of these limitations to future work, (3) Since NSG aligns the video with the entire task graph, it requires the full task execution video. Without this, alignment is partial, rendering NSG ineffective for online task verification.

%%%%%%%%%%%%%%%%%%%%%%%%%%%%%%%%%%%%%%%%%%%%%%%%%%%%%%%%%%%%%%%%%%%%%%%%%%%%%%%%%%%t

%% file: figures/complexity-confusion_mat.tex
\begin{figure}[t]
\centering
    \includegraphics[width=\linewidth]{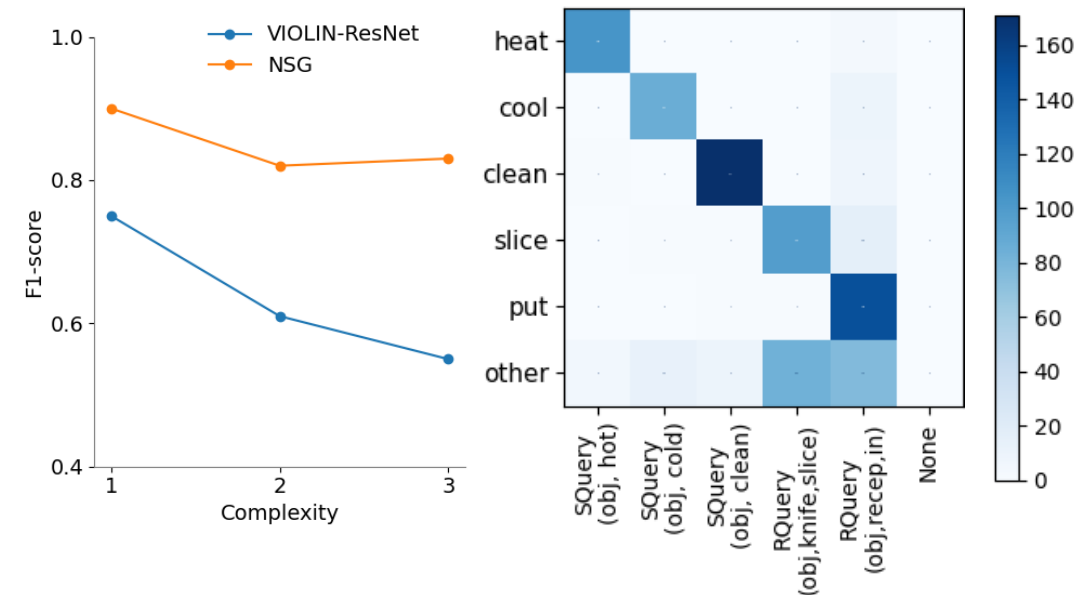}
    \caption{[Left] F1-score of NSG vs. best-performing baseline for \etv tasks with varying complexity averaged over all splits (Appendix~\ref{appendix:analysis} shows performance with varying ordering). [Right] Confusion Matrix for NSG Queries on validation split (SQuery: \code{StateQuery}, RQuery: \code{RelationQuery}). See Appendix~\ref{appendix:analysis} for results on all splits.}
    \label{figure:complexity-confusion_mat}
    % \vspace{-0.1cm}
\end{figure}

%% file: sections_CRV/6_Conclusion.tex
\section{Conclusion}
To address various challenges towards the development of egocentric assistants that can track and verify the accomplishment of everyday tasks, including reasoning about causal and temporal constraints in tasks, visual grounding, and compositional generalization, we introduce Egocentric Task Verification (EgoTV), a benchmark and dataset containing partially-ordered, multi-step tasks with natural language task specifications. We also present NSG, a novel neuro-symbolic approach that enables order-aware visual grounding, and demonstrate its effectiveness on both the EgoTV dataset and a real-world dataset CTV. We hope our contributions will help in advancing research on egocentric assistants that can aid users in everyday tasks.

%%%%%%%%%%%%%%%%%%%%%%%%%%%%%%%%%%%%%%%%%%%%%%%%%%%%%%%%%%%%%%

%% file: sections_CRV/7_Acknowledgement.tex
\section*{Acknowledgement}
This work was partially supported by the Wallenberg AI, Autonomous Systems and Software Program (WASP) funded by the Knut and Alice Wallenberg Foundation.

%% file: sections_CRV/sup.tex
\newpage

\twocolumn[
\begin{center}
\textbf{\Large Appendix for ``\etv \inlineimg{figures/TV}: Egocentric Task Verification \\from Natural Language Task Descriptions"}
\end{center}
\hfill \break
\hfill \break
]

\noindent \textbf{This appendix is organized as follows:} \\
\noindent 8. \emph{EgoTV dataset}: Generation details, additional statistics. 
9. \emph{CrossTask Verification (CTV) dataset}: Details on generation process and evaluation. \\
10. \emph{NSG}: Semantic parsing approaches, additional analysis and ablations, performance on CTV. \\
11. \emph{Baselines}: Details on VLM and Oracle baselines.

%\noindent 11. EgoTV dataset tables and charts 

% We will release the code and dataset upon acceptance. A few sample videos \& annotations from the \etv dataset are provided in {\color{blue}\textbf{EgoTV data/}} in the supplementary material. We also include demo videos showing how NSG works on the dataset in the folder {\color{blue}\textbf{DemoVideo/}}. Lastly, the code for generating the dataset as well as for NSG and the baselines is in the folder {\color{blue}\textbf{Code/}}. 

\input{supplement/7_Ego_Dataset}
\input{supplement/8_CTV_dataset}
\input{supplement/9_NSG_details}

\input{supplement/10_Baseline_details}
%\input{supplement/11_EgoTV_tables_charts}

% \paragraph{List of tasks:}

%% file: supplement/7_Ego_Dataset.tex
\section{\etv Dataset}
\subsection{Task-video Generation using PDDL Planner}
\label{appendix:etv_taskvideogen}
To generate \etv tasks, we encode the final state of the objects achieved by an \etv task as the ``goal state" for the Planning Domain Definition Language (PDDL) planner. Note that the ordering constraints of the tasks aren't captured when the tasks are encoded as goal states for the PDDL planner in this manner. For instance, the tasks of~\emph{clean\_then\_heat(apple)} and ~\emph{clean\_and\_heat(apple)} would have the same PDDL goal states. Consequently, we enforce ordering constraints for a given task using ``pre-conditions" in PDDL. In the above example task of~\emph{clean\_then\_heat(apple)}, the \emph{clean} sub-task would thus be the pre-condition for the \emph{heat} sub-task. Apart from the tasks and object states, the agent state and environment dynamics -- e.g., the \emph{heat} sub-task changes the state of the target object to be hot -- are also encoded in PDDL. 

For each task, a random kitchen scene is picked and the agent is spawned in the corresponding AI2-THOR scene. The planner leverages the initial state and action definitions to generate a sequence of sub-tasks required to achieve the goal. Rather than simply selecting the best plan, which may not always reflect human-like behavior, we aim to mimic the less-than-optimal decision-making of humans by randomly selecting a plan from the top-$k$ plans. This approach enables the inclusion of sub-tasks that may not be strictly necessary for achieving the goal. Furthermore, the partial-ordered nature of tasks enables different plan generations to achieve the same task, thus promoting diversity. We use the Fast Forward (FF) planner~\cite{metric_ff} to generate plans. 

Apart from sequencing the sub-tasks appropriately for achieving a given task, the planner also ensures that the agent can navigate to the correct locations for each sub-task. For instance, to execute the~\emph{clean} sub-task, an agent typically requires using the sink in the kitchen and hence must navigate there. The generated plans thus consist of navigation and object interaction actions.

% the initial state of the agent is obtained by randomly spawning it in an agent is randomly spawned in an AI2-THOR scene for each \etv task. The planner computes a sequence of sub-tasks, which renders the initial state of the agent to the desired final state for each task. Apart from sequencing the sub-tasks appropriately for achieving a given task, the planner also ensures that the agent can navigate to the correct locations for each sub-task. For instance, to execute the~\emph{clean} sub-task, an agent typically requires using the sink in the kitchen and hence must navigate there. The generated plans thus consist of navigation and object interaction actions.

% We further leverage the partial-ordered nature of tasks in our dataset to generate multiple, valid plans for a given task instead of a single, best plan. For instance, for the task of~\emph{clean\_and\_heat\_then\_slice(apple)}, we generate two plans and for the task of~\emph{clean\_and\_heat\_and\_slice(apple)}, we generate six plans corresponding to all the valid permutations. 

\subsection{\etv Task-description Generation}
\label{appendix:task_templates}
% Each task in \etv is associated with $\approx$10 different templates (inclusive of positive and negative scenarios). We also keep an additional template set for the abstraction split. 
The task descriptions corresponding to negative samples, where the task videos are not entailed by their descriptions, are created by either altering the sequence of sub-tasks in the positive template or by replacing some of them with alternative sub-tasks picked randomly from the remaining repertoire of sub-tasks (see Figure~\ref{figure:dataset} where \textit{heat} is replaced with \textit{cool}).  To ensure the practicality of an assistive agent that aids a human, we maintain the target object in the negative samples but vary the sub-tasks, as negative task descriptions on the sub-task level are more relevant than on the object level. For abstraction, we (i) omit the low-level details like \textit{clean \underline{in the sinkbasin}, cool \underline{in the fridge}}; (ii) (some) task-oriented descriptions are changed to goal-oriented descriptions (\textit{apple is heated and cleaned} $\mapsto$ \textit{hot, clean apple}). 

An example template of task descriptions corresponding to the task \emph{cool\_then\_clean} is [`\{obj\} is cooled in a Fridge, then cleaned in a SinkBasin',  `\{obj\} is cleaned in a SinkBasin after cooling in a Fridge', `\{obj\} is cooled in a Fridge before cleaning in a SinkBasin']. While \etv already incorporates some diversity in task descriptions in this manner, we note that inclusion of more diverse free-form language descriptions in the dataset (possibly collected through crowdsourcing) would be a valuable future enhancement.

\subsection{Dataset Analysis and Statistics}
\label{appendix:dataset_analysis_and_statistics}
See Figure~\ref{figure:video-text-analysis} for a comparison of video lengths and task description lengths across different splits. It can be observed that the Novel Tasks split has the longest videos ($\approx 1.6$ minutes) and task descriptions ($\approx 12$ words) owing to its compositional tasks. Additionally, the Abstraction split has the shortest task description ($\approx 5$ words), even for longer videos due to abstraction. We include all tasks of \etv in Table~\ref{tab:list_of_tasks} at the end of the supplement. We also perform a detailed analysis of each split (Figure~\ref{figure:tasks-all-splits}).
% We also perform a detailed analysis of each split as shown in Figures~\ref{figure:train_analysis}~\ref{figure:novel_tasks_analysis}~\ref{figure:novel_steps_analysis}~\ref{figure:novel_scenes_analysis}~\ref{figure:abstraction_analysis}. 

\input{tables/Tab_1_full}
\input{figures/sup/video-text_analysis}
% \input{figures/sup/extra-analysis}

%% file: tables/Tab_1_full.tex
\begin{table*}[t]
\centering
% \footnotesize
\footnotesize{
\begin{tabular}{lcccccccc}
% \hline
& \multicolumn{2}{c}{\textbf{-------- Reasoning --------}} & \multicolumn{4}{c}{\textbf{-----------  Dataset Characteristics -----------}} & \multicolumn{2}{c}{\textbf{--- Grounding ---}} \\
 & \begin{tabular}[c]{@{}c@{}}compos-\\ itional\end{tabular} & causal & language & egocentric & \begin{tabular}[c]{@{}c@{}}real-\\ world\end{tabular} & \begin{tabular}[c]{@{}c@{}}diagnostic\\ tools\end{tabular} & \begin{tabular}[c]{@{}c@{}}objects,\\ relations\end{tabular} & actions\\
\hline
% CLEVR~\cite{clevr_dataset} & \multicolumn{3}{l}{\qquad \cmark $\kern 2.3pc$ \xmark $\kern 2.7pc$ \xmark} & \multicolumn{5}{l}{\qquad \cmark $\kern 2.7pc$ \xmark $\kern 2.8pc$ \xmark $\kern 2.8pc$ \xmark $\kern 3.2pc$ \cmark} & \multicolumn{2}{l}{\qquad \cmark $\kern 2.1pc$ \xmark} \\ 
% % \hline 
% GQA~\cite{gqa_dataset} & \multicolumn{3}{l}{\qquad \cmark $\kern 2.3pc$ \xmark $\kern 2.7pc$ \xmark} & \multicolumn{5}{l}{\qquad \cmark $\kern 2.7pc$ \xmark $\kern 2.8pc$ \cmark $\kern 2.8pc$ \xmark $\kern 3.2pc$ \cmark} & \multicolumn{2}{l}{\qquad \cmark $\kern 2.1pc$ \xmark} \\ 
% % \hline
% \begin{tabular}[c]{@{}l@{}}Visual \\ Genome~\cite{visual_genome}\end{tabular}  & \multicolumn{3}{l}{\qquad \cmark $\kern 2.3pc$ \xmark $\kern 2.7pc$ \xmark} & \multicolumn{5}{l}{\qquad \cmark $\kern 2.7pc$ \xmark $\kern 2.8pc$ \cmark $\kern 2.8pc$ \xmark $\kern 3.2pc$ \xmark} & \multicolumn{2}{l}{\qquad \cmark $\kern 2.1pc$ \xmark} \\ 
% \hline
CLEVRER~\cite{clevrer} & \multicolumn{2}{l|}{\qquad \cmark $\kern 3.5pc$ \cmark} & \multicolumn{4}{l|}{\qquad \cmark $\kern 2.7pc$ \xmark $\kern 2.8pc$ \xmark $\kern 3.2pc$ \cmark} & \multicolumn{2}{l}{\qquad \cmark $\kern 2.1pc$ \xmark}  \\ 
% \hline
Next-QA~\cite{next_qa_dataset} & \multicolumn{2}{l|}{\qquad \xmark $\kern 3.5pc$ \cmark} & \multicolumn{4}{l|}{\qquad \cmark $\kern 2.7pc$ \xmark $\kern 2.8pc$ \cmark $\kern 3.2pc$ \cmark} & \multicolumn{2}{l}{\qquad \cmark $\kern 2.1pc$ \cmark} \\ 
% \hline
AGQA~\cite{agqa_dataset} & \multicolumn{2}{l|}{\qquad \cmark $\kern 3.5pc$ \xmark} & \multicolumn{4}{l|}{\qquad \cmark $\kern 2.7pc$ \xmark $\kern 2.8pc$ \cmark $\kern 3.2pc$ \cmark} & \multicolumn{2}{l}{\qquad \cmark $\kern 2.1pc$ \cmark} \\ 
% \hline
\begin{tabular}[c]{@{}l@{}}Activity\\ Net-QA~\cite{activity_net_dataset}\end{tabular} & \multicolumn{2}{l|}{\qquad \xmark $\kern 3.5pc$ \xmark} & \multicolumn{4}{l|}{\qquad \cmark $\kern 2.7pc$ \xmark $\kern 2.8pc$ \cmark $\kern 3.2pc$ \xmark} & \multicolumn{2}{l}{\qquad \cmark $\kern 2.1pc$ \cmark}\\ 
% \hline
STAR~\cite{star_situated_reasoning} & \multicolumn{2}{l|}{\qquad \cmark $\kern 3.5pc$ \cmark} & \multicolumn{4}{l|}{\qquad \cmark $\kern 2.7pc$ \xmark $\kern 2.8pc$ \cmark $\kern 3.2pc$ \cmark} & \multicolumn{2}{l}{\qquad \cmark $\kern 2.1pc$ \cmark}\\ 
% \hline
CoPhy~\cite{cophy_dataset} & \multicolumn{2}{l|}{\qquad \xmark $\kern 3.5pc$ \cmark} & \multicolumn{4}{l|}{\qquad \xmark $\kern 2.8pc$ \xmark $\kern 2.9pc$ \xmark $\kern 3.2pc$ \cmark} & \multicolumn{2}{l}{\qquad \cmark $\kern 2.1pc$ \xmark} \\ 
% \hline
Social-IQ~\cite{social_iq} & \multicolumn{2}{l|}{\qquad \xmark $\kern 3.5pc$ \cmark} & \multicolumn{4}{l|}{\qquad \cmark $\kern 2.7pc$ \xmark $\kern 2.8pc$ \cmark $\kern 3.2pc$ \cmark} & \multicolumn{2}{l}{\qquad \xmark $\kern 2.1pc$ \xmark} \\ 
% \hline
\begin{tabular}[c]{@{}l@{}}Causal-\\ VidQA~\cite{causal_vid_qa}\end{tabular} & \multicolumn{2}{l|}{\qquad \xmark $\kern 3.5pc$ \cmark} & \multicolumn{4}{l|}{\qquad \cmark $\kern 2.7pc$ \xmark $\kern 2.8pc$ \cmark $\kern 3.2pc$ \cmark} & \multicolumn{2}{l}{\qquad \cmark $\kern 2.1pc$ \cmark}\\ 
% \hline
Charades~\cite{charades_dataset} & \multicolumn{2}{l|}{\qquad \xmark $\kern 3.5pc$ \xmark} & \multicolumn{4}{l|}{\qquad \cmark $\kern 2.7pc$ \xmark $\kern 2.8pc$ \cmark $\kern 3.2pc$ \cmark} & \multicolumn{2}{l}{\qquad \cmark $\kern 2.1pc$ \cmark} \\ 
% \hline
% \begin{tabular}[c]{@{}l@{}}Action \\ Genome~\cite{action_genome}\end{tabular} & \multicolumn{3}{l|}{\qquad \xmark $\kern 2.3pc$ \xmark $\kern 2.7pc$ \xmark} & \multicolumn{5}{l|}{\qquad \xmark $\kern 2.7pc$ \cmark $\kern 2.8pc$ \cmark $\kern 2.8pc$ \cmark $\kern 3.2pc$ \xmark} & \multicolumn{2}{l}{\qquad \cmark $\kern 2.1pc$ \cmark}\\ 
% \hline
CATER~\cite{cater_dataset} & \multicolumn{2}{l|}{\qquad \cmark $\kern 3.5pc$ \xmark} & \multicolumn{4}{l|}{\qquad \xmark $\kern 2.8pc$ \xmark $\kern 2.9pc$ \xmark $\kern 3.2pc$ \cmark} & \multicolumn{2}{l}{\qquad \cmark $\kern 2.1pc$ \cmark}\\
% \hline
\begin{tabular}[c]{@{}l@{}}EPIC-\\ KITCHENS~\cite{epic_kitchens}\end{tabular} & \multicolumn{2}{l|}{\qquad \cmark $\kern 3.5pc$ \cmark} & \multicolumn{4}{l|}{\qquad \cmark $\kern 2.7pc$ \cmark $\kern 2.7pc$ \cmark $\kern 3.2pc$ \cmark} & \multicolumn{2}{l}{\qquad \cmark $\kern 2.1pc$ \cmark}\\
% \hline
Ego-4D~\cite{ego_4d} & \multicolumn{2}{l|}{\qquad \xmark $\kern 3.5pc$ \cmark} & \multicolumn{4}{l|}{\qquad \cmark $\kern 2.7pc$ \cmark $\kern 2.7pc$ \cmark $\kern 3.2pc$ \cmark} & \multicolumn{2}{l}{\qquad \cmark $\kern 2.1pc$ \cmark}\\ 
% \hline
VIOLIN~\cite{violin_dataset} & \multicolumn{2}{l|}{\qquad \xmark $\kern 3.5pc$ \cmark} & \multicolumn{4}{l|}{\qquad \cmark $\kern 2.7pc$ \xmark $\kern 2.8pc$ \cmark $\kern 3.2pc$ \xmark} & \multicolumn{2}{l}{\qquad \cmark $\kern 2.1pc$ \cmark}\\
% \hline
Change-It~\cite{change_it} & \multicolumn{2}{l|}{\qquad \xmark $\kern 3.5pc$ \cmark} & \multicolumn{4}{l|}{\qquad \xmark $\kern 2.7pc$ \cmark $\kern 2.8pc$ \cmark $\kern 3.2pc$ \xmark} & \multicolumn{2}{l}{\qquad \cmark $\kern 2.1pc$ \cmark}\\
% \hline
Cross-Task~\cite{cross_task} & \multicolumn{2}{l|}{\qquad \cmark $\kern 3.5pc$ \cmark} & \multicolumn{4}{l|}{\qquad \cmark $\kern 2.7pc$ \xmark $\kern 2.8pc$ \cmark $\kern 3.2pc$ \xmark} & \multicolumn{2}{l}{\qquad \cmark $\kern 2.1pc$ \cmark}\\
\hline
\textbf{EgoTV} \parbox[c]{1em}{
      \includegraphics[width=0.1in]{figures/TV}} & \multicolumn{2}{l|}{\qquad \blueD{\cmark} $\kern 3.3pc$ \blueD{\cmark}} & \multicolumn{4}{l|}{\qquad \blueD{\cmark} $\kern 2.5pc$ \blueD{\cmark} $\kern 2.5pc$ \red{\xmark} $\kern 3.0pc$ \blueD{\cmark}} & \multicolumn{2}{l}{\qquad \blueD{\cmark} $\kern 1.9pc$ \blueD{\cmark}}\\
\hline
\end{tabular}
}
\vspace{2mm}
\caption{\textbf{\etv vs. existing video-language datasets.} \etv benchmark enables reasoning (compositional, causal, and temporal); has unique dataset characteristics along with diagnostics; requires grounding of objects, relations, and actions.}
\label{table:list_of_datasets_full}
\end{table*}

%% file: figures/sup/video-text_analysis.tex
\begin{figure}[ht]
\centering
    \includegraphics[width=\linewidth]{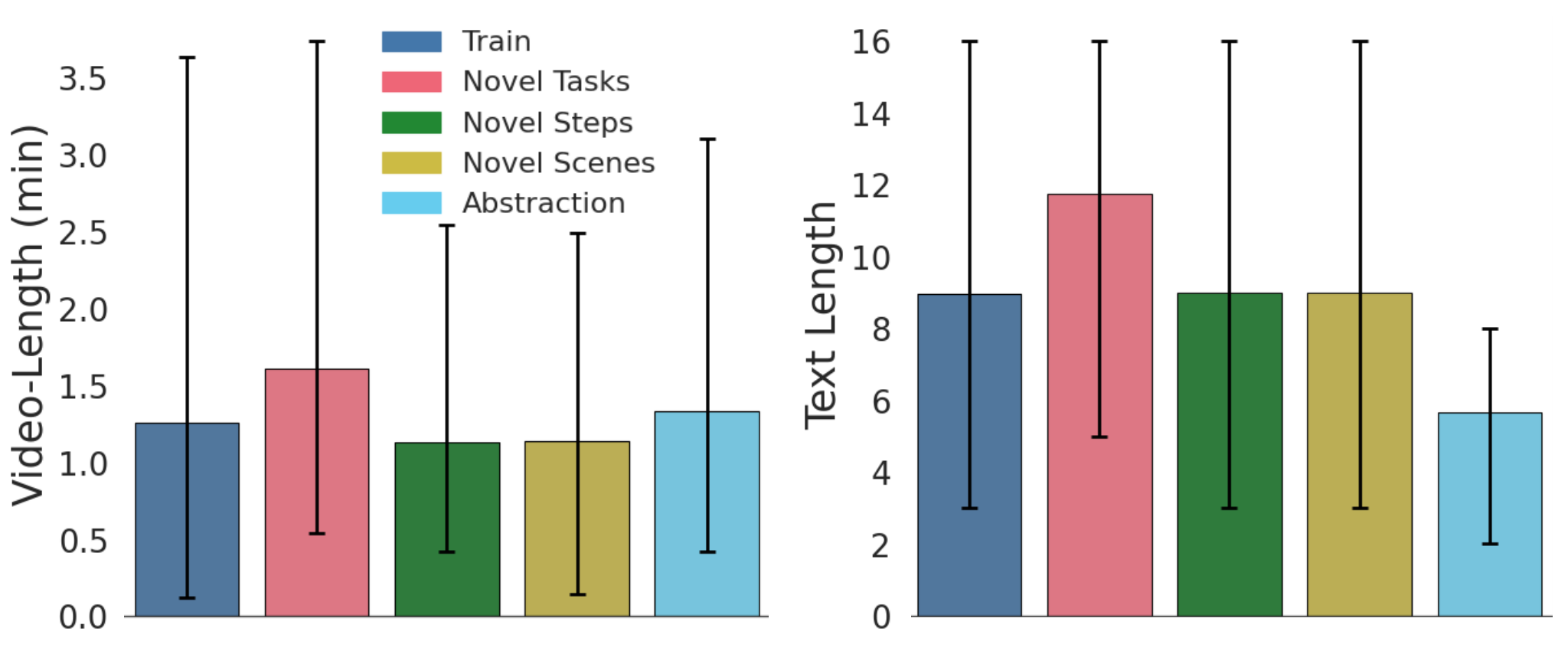}
    \caption{\textbf{Additional \etv dataset statistics.} Comparison of video length (in minutes) and task description length (number of words) across different splits using error plots. Novel Tasks split has the longest videos and task descriptions. Abstraction split has the shortest task descriptions.}
    \label{figure:video-text-analysis}
% \vspace{-0.3cm}
\end{figure}

%% file: supplement/8_CTV_dataset.tex
%%%%%%%%%%%%%%%%%%%%%%%%%%%%%%%%%%%%%%%%%%%%%%%%%%%%%%%%%%%%%%%%%%%%%
%%%%%%%%%%%%%%%%%%%%below%%%%%%%%%%%%%%%%%%%%%%%%%%%%%%%
%%%%%%%%%%%%%%%%%%%%%%%%%%%%%%%%%%%%%%%%%%%%%%%%%%%
%\input{tables/task_topics}
%%%%%%%%%%%%%%%%%%%%%%%%%%%%%%%%%%%%%%%%%%%%%%%%%%%%%%%%%%%%%%%%%%%%%
%%%%%%%%%%%%%%%%%%%above%%%%%%%%%%%%%%%%%%%%%%%%%%%%%%%%
%%%%%%%%%%%%%%%%%%%%%%%%%%%%%%%%%%%%%%%%%%%%%%%%%%%

\section{CrossTask Verification (CTV) Dataset}
\label{appendix:cross_task_construct}
\input{figures/task_graph}

\subsection{Dataset construction}
\label{appendix:ctv_dataset_construction}
We leverage the CrossTask dataset and its action step annotations to construct our CrossTask Verification dataset. % with the following two settings: 
%\noindent (1) \textbf{Action sequence verification.} 
Each video contains task descriptions that are obtained by concatenating the sequence of \textit{action steps} annotations available in the original CrossTask dataset. 
For ease of experimentation, we only consider the top-4 frequent action steps for each task in CrossTask. Consequently, CTV dataset videos are constructed by selecting the segments corresponding to these frequent actions per video and are thus shorter than the original CrossTask videos.

%To ensure repetitive subtasks, as in \etv, we start by selecting the action steps within each task based on frequency. Then, we select the four most frequent action steps for each task and filter out video segments in CrossTask that do not contain these action steps.

% for each video, as our task description.
% \begin{itemize}[leftmargin=*,noitemsep]
%     \item 

    % \item 
    We follow a process similar to EgoTV for generating \textbf{negative descriptions}: (i) \textit{replacing action steps from other tasks}: where we substitute an action step in the sequence with an action step in another task. (ii) \textit{replacing action steps from the same task}: instead of replacing steps from another task, we reuse the unused action steps, which were not the top-4 frequent steps in the same task. These action steps are closer in semantics and serve as hard negative. (iii) \textit{replacing action step sequence order to an impossible order}: we first list out all possible action step orders in CrossTask following \cite{mao2023action} as shown in Figure \ref{figure:cross_task_verification_graph}. Then, we construct an action step sequence that doesn't exist in CrossTask for the given task, e.g., \textit{lower jack, upper jack, break on} for the task of \emph{fixing the car}. We assume that these sequence orders are impossible since they were not observed in any videos for a given task.
    
    Apart from generating negative NL task descriptions, we also generate \textbf{negative videos} in two ways:~(i) \textit{shuffling the video order corresponding to action step sequence}: this mimics the case where the action steps were not executed in the correct order. (ii) \textit{dropping a video segment corresponding to an action step}: this corresponds to the situations where an action step is missed while executing the task, potentially resulting in a failure in accomplishing the task. Thus, this is an important use case for task verification.

%% file: figures/task_graph.tex
\begin{figure}[ht]
\centering
    \includegraphics[width=0.9\linewidth]{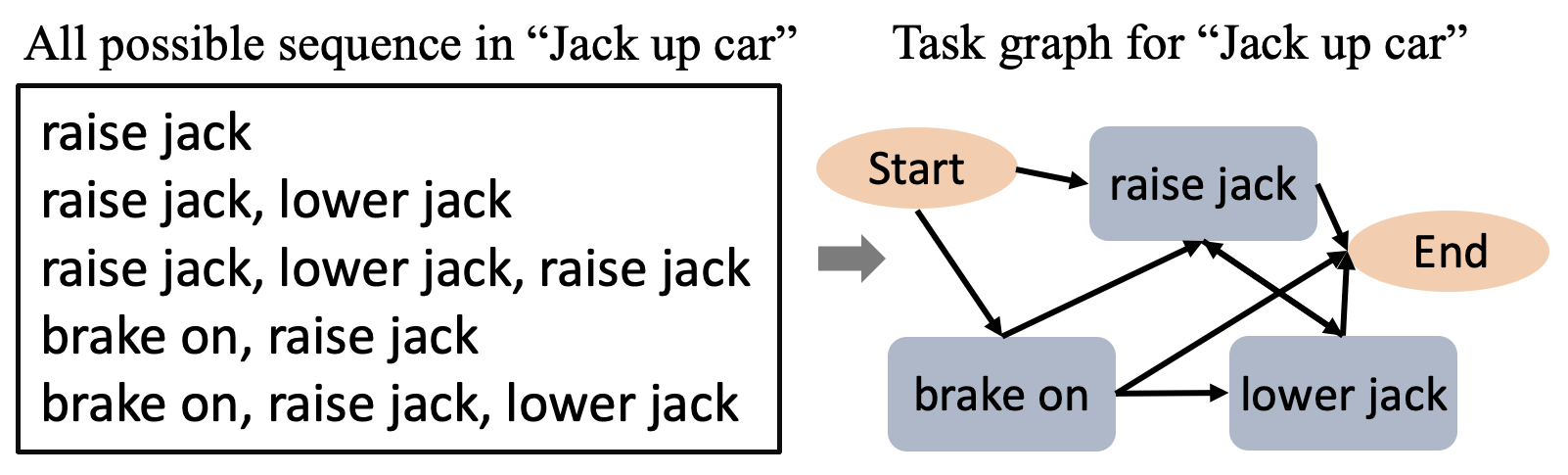}
    \vspace{10pt}
    \caption{\textbf{Task graphs are used to create impossible sequences for generating negative task descriptions in the CTV dataset.}}
    \label{figure:cross_task_verification_graph}
%\vspace{-20pt}
\end{figure}

%% file: supplement/9_NSG_details.tex
\section{NSG details}

%%%%%%%%%%%%%%%%%%%%%%%%%%%%%%%%%
% The query types used by NSG for both the \etv dataset and the CTV dataset are shown in Table~\ref{table:QTypes}.

% \subsection{NSG code}
% We include our NSG code in the folder {\color{blue}\textbf{Code/nsg}} comprising the training code for \code{(State+Relation)Query} model in \textit{nesy\_v1.py} and \code{ActionQuery} model in \textit{nesy\_v2.py}.

%%%%%%%%%%%%%%%%%%%%%%%%%%%%%%
\subsection{NSG Training details}
\label{appendix:nsg_training}
When training NSG, we do not update the weights of the CLIP feature extractors (Sec.~\ref{subsection:query_encoders}) due to GPU memory limitations. We use a batch of N = 64 samples, where we sample the video at 2.5 FPS. We set a window size $k=20$ frames for segmentation in NSG (Sec.~\ref{subsection:plan_verification}), each window representing an 8-second video segment. We use a train-validation split of 80-20 and use the validation performance as an indicator of convergence. We minimize the binary cross-entropy loss in Eq.~\ref{eq:bce} with Adam \cite{kingma2014adam} and a learning rate of 1e-3. Each model is trained on 8 V100 GPUs for 50 epochs for two days.
% \subsection{Graph generation for NSG}
% We follow AAAI workshop to create graph by frequency

%%%%%%%%%%%%%%%%%%%%%%%%%%%%%%%%
\subsection{NSG Semantic Parsing}
\label{section:appendix_semantic_parsing}
% Given, two NL task descriptions: (A) \textit{apple is heated, then cleaned in a sink basin}; (B) \textit{hot, clean apple}, one can infer that description (A) requires the apple to be heated first and then cleaned, whereas, in description (B) the apple must be heated and cleaned, but the order of heating and cleaning is not specified.
We test two semantic parsing methods: (i) Finetuning language models to generate graphs from NL descriptions; (ii) Few-shot prompting of large language models.
%%%%%%%%%%%%%%%%%%%%%%%%%%%%%%%%%%%%%
\subsubsection{Finetuning Language Models}
Recently, it has been shown that pre-trained language models can be leveraged for graph~\cite{proscript} and plan generation~\cite{llms_for_translating_to_goal} from NL. We use a similar framework to train a T5-small transformer~\cite{tf_transformer} to generate partial-ordered plans. For this, we use a subset of the training data (in particular the positive task descriptions for which we have gold-label graphs annotations) and annotate them with their partial-ordered plans in the form of directed acyclic graphs (DAG) -- $G(V,E)$, where vertices $n_i \in V$ represent sub-tasks, and edges $e_{ij} \in E$ are ordering constraints that indicate $n_i$ must precede $n_j$ (i.e. $n_i \rightarrow n_j$). To train the text generation transformer, we represent the output graph in DOT language. The corresponding DOT representation for the graph in Figure~\ref{figure:model-layout} is given as: \code{Step~0:~StateQuery(apple,hot), Step~1:~StateQuery(apple,clean), Step~2:~StateQuery(apple,sliced), Step~3:~RelationQuery(apple,plate,in), Step~0~$\rightarrow$~Step~1, Step~0~$\rightarrow$~Step~2, Step~1~$\rightarrow$~Step~3, Step~2~$\rightarrow$~Step~3} \\

\textbf{Ablations on Plan Generation framework}: We assess the correctness of the graphs generated by the trained T5-transformer model on the positive task descriptions of our test splits. The evaluation metric is Graph Edit Distance (GED)~\cite{ged} which computes the distance between two graphs ($G_1$ and $G_2$) given as:
    \begin{equation*}
        GED(G_1, G_2) = \min_{G_1 \xrightarrow{d_1, \dots, d_k} G_2} \sum_{i=1}^k cost(d_i)
    \end{equation*}
where, $d_1, \dots d_k$ are graph edit operations (insertion, deletion, replacement of a vertex or an edge) from $G_1$ to $G_2$. With the exception of the abstraction split\footnote{For abstraction, we observed syntax errors, e.g., missing/incorrect arguments for queries in the generated graphs like \code{RelationQuery(apple, slice)}, instead of \code{RelationQuery(apple, \underline{knife}, slice)}. However, the selection of the correct query module and the partial grounding of the correct arguments lead to a significant improvement over baselines.}, the GED for all test splits was observed to be $\approx 0.03$ (GED $\downarrow$). Here, $\downarrow$ signifies that a lower GED score is better, with the lowest value being $=0$. Note, that although it is possible for the response to contain syntax errors, e.g., invalid names or queries with invalid arguments, the output can be improved by leveraging the DSL grammar to avoid invalid arguments.

%%%%%%%%%%%%%%%%%%%%%%%%%%%%%%%%%%%%%%%%%%5
\subsubsection{Prompting Language Models} 
We also experimented with prompting. As a proof-of-concept, we show our example prompts in Table~\ref{prompting_box}. We use ChatGPT~\cite{chatGPT} with few-shot prompting. The prompt is displayed in gray, the queries in blue, and the generated output in green. We observed that the use of Chain-of-Thought~\cite{chain-of-thought} improved the output.

%%%%%%%%%%%%%%%%%%%%%%%%%%%%%%%%%%%%%%
\subsection{Integer Programming for Alignment in NSG}
\label{appendix:integer_programming}
The constrained optimization problem defined in Eqs.~\ref{Z-main}, without the ordering constraint, can also be formulated as an Integer Programming problem~\cite{integer_programming} where variables $Z_{jt}$ can only take integer values in $\{0,1\}$ (i.e. with the additional constraint $\mathrm{Z} \in \{0,1\}^{N \times S}$). Notably, the proposed DP solution adheres to the constraints Eqs.~\ref{Z-a} \ref{Z-b} \ref{Z-c}. The first part of the DP (highlighted in green in Figure~\ref{figure:model-layout}) pairs the current query $q_j$ with the current segment $s_t$, then tries to align the rest of the queries with the remaining segments to meet the requirement of one query per segment, as specified in Eq.~\ref{Z-a}. The second part of the solution (denoted by the red box) skips over the current segment $s_t$ and tries to align the same queries with the remaining segments until a pairing is found (i.e. $Z_{jt}=1$). Thus, together with the base case of $\mathrm{Z}=\mathds{I} \; \text{if} \; N=S$, it satisfies Eq.~\ref{Z-b}. Furthermore, since the DP processes the queries and segments in a specific order (topological sorting for queries and temporal order of video segments), it also meets the ordering constraint requirement specified in Eq.~\ref{Z-c}.

\subsection{NSG Analysis}
\label{appendix:analysis}

\begin{itemize}[leftmargin=*,noitemsep]
    \item Table~\ref{table:window_size} demonstrates the robustness of NSG to changes in the window size $k$. 
    
    \item We trained two different NSG models to compare the impact of query types with (i) \code{(State+Relation)Query}, (ii) \code{ActionQuery}. Figure~\ref{figure:all_cf} reveals that a combination of \code{(State+Relation)Query} is effective at detecting sub-tasks in segments with high recall, particularly for \emph{slice} and \emph{put}. We also note that the NSG model struggles to detect the \emph{slice} sub-task in the Novel Steps split.

\end{itemize}

%%%%%%%%%%%%%%%%%%%%%%%%%%%%%%%%%%%%%%%%%%%%%%%%%%%%%%%%%%%%%%%%%%%%%%%%%%

\twocolumn[{\begin{tcolorbox}[fontupper=\small, fontlower=\Large, label=prompting_box, title=Table~\ref{prompting_box}: Semantic Parsing by Prompting ChatGPT]
% \begin{multicols}{2}
\code{Q. slice of apple is heated in microwave, then placed in a plate}\\
\code{A. First the apple is sliced, hence Step 1: slice apple. Then the apple is heated, hence Step 2: heat apple. Then the apple is placed in a plate, hence Step 3: place apple in plate. Hence, the order of the steps are: slice apple < heat apple < place apple in plate}\\

\code{Q. potato is cleaned in sinkbasin and sliced, then cooled in a fridge}\\
\code{A. First the potato is cleaned and sliced in any order, hence Step 1: clean potato; Step 2: slice potato. Then the potato is cooled in a Fridge, hence Step 3: cool potato. Hence, the order of the steps are: clean potato and slice potato < cool potato}\\

\code{Q. hot, sliced, clean tomato}\\
\code{A. First the tomato is heated and sliced and cleaned in any order, hence Step 1: heat tomato; Step 2: slice tomato; Step 3: clean tomato. Hence, the order of the steps are: heat tomato and slice tomato and clean tomato}\\

\query{Q. apple is heated and cleaned in sinkbasin, then cooled and sliced}\\
\answer{A. First the apple is heated and cleaned in a sink basin in any order, hence Step 1: heat apple; Step 2: clean apple. Then the apple is cooled and sliced in any order, hence Step 3: cool apple; Step 4: slice apple. Hence, the order of the steps are: heat apple and clean apple < cool apple and slice apple.}\\

\query{Q. sliced apple is heated in microwave, then cleaned in a sinkbasin}\\
\answer{A. First the apple is sliced, hence Step 1: slice apple. Then the apple is heated in a microwave, hence Step 2: heat apple. Then the apple is cleaned in a sink basin, hence Step 3: clean apple. Hence, the order of the steps are: slice apple < heat apple < clean apple}\\

\query{Q. apple is heated in microwave after cooling and cleaning}\\
\answer{First the apple is cooled and cleaned in any order, hence Step 1: cool apple; Step 2: clean apple. Then the apple is heated in a microwave, hence Step 3: heat apple. Hence, the order of the steps are: cool apple and clean apple < heat apple.}
% \end{multicols}
\end{tcolorbox}}]

%%%%%%%%%%%%%%%%%%%%%%%%%%%%%%%%%%%%%%%%%%%%%%%%%%%%%%%%%%%%%%%%%%%%%%%%

%%%%%%%%%%%%%%%%%%%%%%%%%%%%%%%%%%%%%%%%%%%%%%%%%%%%%%%%%%

\begin{itemize}[leftmargin=*,noitemsep]
    \item We evaluated the NSG model against the best-performing baseline~(VIOLIN-ResNet)~across two axes:~task complexity and ordering.~Figure~\ref{figure:complexity-ordering} shows that NSG's performance is robust to complexity and ordering variations.
    \item Since NSG only utilizes the features from aligned video segments and ignores the remaining segments, we conducted experiments to investigate the role of context (from the discarded segments) in \etv. Specifically, we trained an extra BiLSTM layer to encode bidirectional context from adjacent segments on top of the CLIP segment features. From Table~\ref{table:window_size}, we observed improved performance across all splits, except for Novel Tasks. We attribute this to a potential loss of compositional and temporal comprehension of the segment features caused by the inclusion of additional context information.
\end{itemize}

\label{appendix:NSG_crosstask}

\subsection{NSG on Real-world Data}

\paragraph{NSG for CTV.}

In Section \ref{subsection:queries}, we described two symbolic operations to process task description. In CTV, all of the action steps refer to a certain action. Hence, we apply the \code{ActionQuery} as in Table \ref{table:QTypes} to encode all action steps. In Section \ref{subsection:plan_parsing_from_instructions}, we described (1) how we processed task description into a graph and (2) how we generated all possible sequences from the graph. %In our \textit{action sequence verification} setting, 
Since the action sequences are already given in CTV, we can skip the above two steps in Section \ref{subsection:plan_parsing_from_instructions} and directly feed the sequence in our Query Encoder and Video Aligner models. %In our \textit{task verification} setting, only the task class was given. 

\input{tables/table_ablations}

\input{figures/complexity-ordering}
\input{figures/sup/all_cf}
\input{tables/crosstask_results}

%%%%%%%%%%%%%%%%%%%%%%%%%%%%%%%%%%%%%%%%%%%%%%%

%Here, instead of generating a graph from the task description, we follow \cite{mao2023action} to construct a graph given each task by all possible sequence order as shown in Figure \ref{figure:cross_task_verification_graph}. Then, we feed the graph to the rest of our model. 

\paragraph{Performance Evaluation.}
To evaluate how NSG enables task verification in the real world, we compare the performance of NSG against selected baselines described in Sec.~\ref{subsection:baseline_evaluation}, which were either applied to the previous CrossTask evaluation (MIL-NCE\cite{miech2020end}, VideoCLIP\cite{videoclip}) or had competitive performance (CoCa\cite{coca}, VIOLIN\cite{violin_dataset}) on \etv. Note that the methods for CrossTask are not directly applicable to CTV since CTV focuses on task verification (predicting if the task is accomplished) instead of temporal localization (localizing action temporally). Table~\ref{table:crosstask_results} shows that the baseline models VideoCLIP, CoCa perform better than VIOLIN on CTV as compared to \etv. This indicates that CTV is potentially able to better harness the gains from the large-scale VL pretraining in these models compared to \etv. Despite these gains, NSG outperforms the baselines by a significant margin. % in both action sequence and task verification settings. %We also found the \textit{Task verification} setting is more challenging than \textit{Action sequence verification} setting in general since the former offers much less information for the model to predict. Also, the same task may have different ways to accomplish it. The model needs to learn such diverse steps inherently. 
%The results demonstrate the applicability of NSG's causal and compositional reasoning capabilities in the real world.

%% file: tables/table_ablations.tex
% \begin{table}[t]
% \small
% \centering
% \begin{tabular}{lcccc}
% \hline
% \multirow{2}{*}{Model} & \begin{tabular}[c]{@{}c@{}} Novel  \end{tabular} & \begin{tabular}[c]{@{}c@{}}Novel \end{tabular} & \begin{tabular}[c]{@{}c@{}}Novel\end{tabular} & \multirow{2}{*}{Abstract.} \\ & Tasks & Steps & Scenes & \\
% \hline
% \code{ActionQuery} & 78.2 & 45.6 & 70.6 & 75.5\\
% \hline
% $k=12$ &90.6 & 50.3 & 81.3 & 82.5\\
% $k=32$ & 89.2 & 54.9 & 81.0 & 82.3\\
% \hline
% Default & 90.0 &  64.7& 84.9 & 80.4\\
% \hline
% \end{tabular}
% \caption{Ablation studies on \etv: NSG with \code{ActionQuery}; NSG with different window sizes ($k=12, k=32$). The best performance for NSG was obtained with $k=20$, as reported in Table~\ref{table:baseline_results}. \bc{we can add sample rate if we want.}}
% \label{table:table_ablations}
% \end{table}  

\begin{table}[t]
\small
\centering
\small{\begin{tabular}{lcccc}
\hline
\multirow{2}{*}{NSG} & \begin{tabular}[c]{@{}c@{}} Novel  \end{tabular} & \begin{tabular}[c]{@{}c@{}}Novel \end{tabular} & \begin{tabular}[c]{@{}c@{}}Novel\end{tabular} & \multirow{2}{*}{Abstraction} \\ & Tasks & Steps & Scenes & \\
% \code{ActionQuery} & 78.2 & 45.6 & 70.6 & 75.5\\
\hline
$k=12$ &90.6 & 50.3 & 81.3 & 82.5\\
$\boldsymbol{k=20}$ (default) & 90.0 &  64.7& 84.9 & 80.4\\
$k=32$ & 89.2 & 54.9 & 81.0 & 82.3\\
\hline
NSG-BiLSTM & 73.3 & 70.2 & 90.0 & 87.2\\
\hline
\end{tabular}}
\vspace{10pt}
\caption{\textbf{NSG ablations.} [Block 1]: NSG with different window sizes $k=\{12,20,32\}$. The best-performing NSG model with $k=20$ is reported in Table~\ref{table:baseline_results}. [Block 2]: NSG with BiLSTM to encode additional context from neighboring segments.}
\label{table:window_size}
%\vspace{-10pt}
\end{table}

% \begin{table}[t]
% \small
% \centering
% \begin{tabular}{lcccc}
% \hline
% \multirow{2}{*}{NSG} & \begin{tabular}[c]{@{}c@{}} Novel  \end{tabular} & \begin{tabular}[c]{@{}c@{}}Novel \end{tabular} & \begin{tabular}[c]{@{}c@{}}Novel\end{tabular} & \multirow{2}{*}{Abstraction} \\ & Tasks & Steps & Scenes & \\
% \hline
% \code{Action} & 78.2 & 45.6 & 70.6 & 75.5\\
% \code{State+Relation} & 90.0 & 64.7 & 84.9 & 80.4\\
% \hline
% \end{tabular}
% \caption{\code{StateQuery} + \code{RelationQuery} vs. \code{ActionQuery}}
% \label{table:query_comparison}
% \end{table}  

%% file: figures/complexity-ordering.tex
% \input{figures/complexity-ordering}

\begin{figure}[t]
\centering
    \includegraphics[width=\linewidth]{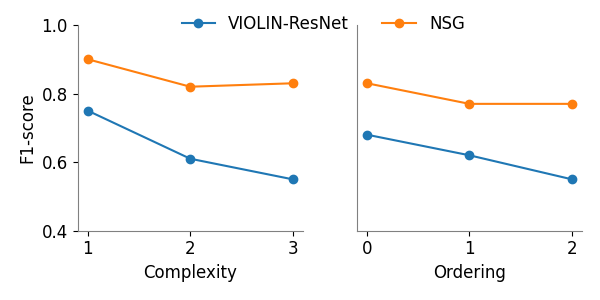}
    \caption{\textbf{NSG maintains consistent performance as task complexity and ordering difficulty increases.} F1-score of NSG vs. best-performing baseline for \etv tasks with varying complexity and ordering are shown.}
    \label{figure:complexity-ordering}
\end{figure}

%% file: figures/sup/all_cf.tex
\begin{figure*}
\centering
    \includegraphics[width=.83\linewidth]{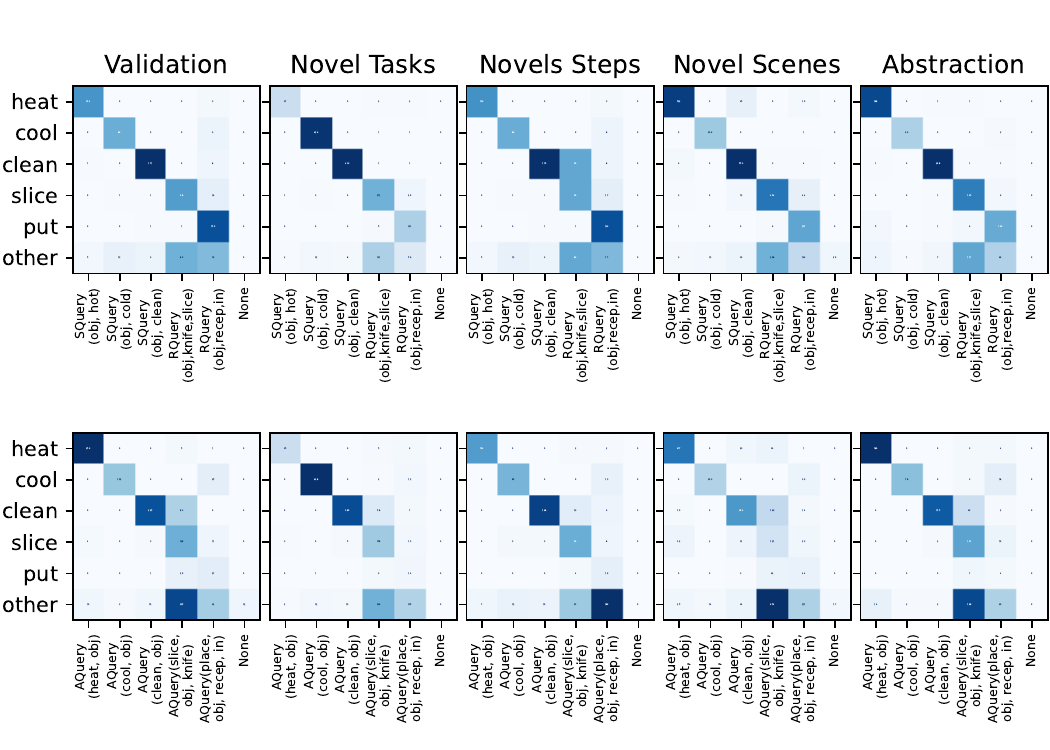}
    %\vspace{-10pt}
    \caption{\textbf{Effect of query types on NSG performance.} Confusion matrices for two different query models across train/test splits of \etv. The Y-axis represents the ground-truth sub-task for a segment, and the X-axis denotes the aligned query for that segment. [Row 1]: Here, SQuery and RQuery denote \code{StateQuery} \& \code{RelationQuery}, respectively. [Row 2]: Here, AQuery denotes \code{ActionQuery}. It can be observed that \code{\{State+Relation\}Query} performs better than \code{ActionQuery}.}
    \label{figure:all_cf}
%\vspace{-10pt}
\end{figure*}

%% file: tables/crosstask_results.tex
% \begin{table}[t]
% \small
% \centering
% \resizebox{\columnwidth}{!}{%
% \begin{tabular}{lcccc}
% \hline
% \textbf{Task desc.} &\textbf{VIOLIN} & \textbf{VideoCLIP} & \textbf{CoCa} & \textbf{NSG}  \\
% \hline
% Action label  & 34.7 & 49.7  & 70.9 & \textbf{76.3}  \\
% Task class  & 32.5 & 43.5  & 65.3 & \textbf{71.4}  \\
% \hline
% \end{tabular}}
% \caption{Performance on CTV's Novel Tasks split}
% \label{table:crosstask_results}
% \end{table}

% \begin{table}[t]
% \small
% \centering
% \begin{tabular}{lcc}
% \hline
% \textbf{Model} &\textbf{Visual Feature} & \textbf{F1} \\
% \hline
% VIOLIN-I3D~\cite{violin_dataset}  & I3D(V) & 34.7 \\
% MIL-NCE~\cite{miech2020end}  & S3D(V) & 43.4 \\
% VideoCLIP~\cite{videoclip} & TX(V) & 49.7 \\
% CoCa \cite{coca}  & TX(I) &  70.9 \\
% \textbf{NSG (ours)} & CLIP(I) & \textbf{76.3} \\
% \hline
% \end{tabular}
% \caption{Performance on CTV's Novel Tasks split}
% \label{table:crosstask_results}
% \end{table}

\begin{table}[t]
\small
\centering
\small{
\begin{tabular}{lcccc}
\hline
\textbf{Model} & Visual & Text&  Fusion & \textbf{F1} \\
\hline
VIOLIN-I3D~\cite{violin_dataset} & I3D & BERT & Y & 34.7 \\
MIL-NCE~\cite{miech2020end} & S3D & Word2Vec &  & 43.4 \\
VideoCLIP~\cite{videoclip} & Tx & Tx & Y & 49.7 \\
CoCa \cite{coca}   & Tx& Tx& Y &  70.9 \\
\textbf{NSG (ours)} & CLIP & CLIP  & Y  & \textbf{76.3} \\
% \hline
% VIOLIN-I3D~\cite{violin_dataset}  & Task class & 32.5 \\
% MIL-NCE~\cite{miech2020end}  & Task class & 40.1 \\
% VideoCLIP~\cite{videoclip} & Task class & 43.5  \\
% CoCa \cite{coca}  & Task class &  65.3 \\
% \textbf{NSG (ours)} & Task class & \textbf{71.4}\\
\hline
\end{tabular}}
\vspace{10pt}
\caption{\textbf{Performance on CTV's test split, which mirrors the Novel Tasks split from \etv}}
% \vspace{-20pt}
\label{table:crosstask_results}
\end{table}

%% file: supplement/10_Baseline_details.tex
\section{Baseline descriptions and details}
\label{appendix:baselines_details}

\subsection{VLM Baselines}
\label{appendix:VLM_baselines}
Majority of VLMs can be characterized based on their approach of extracting and fusing features from vision and text modalities. VLMs for video tasks use either a video encoder or an image encoder with temporal aggregation using sequence model, e.g., transformers~\cite{tf_transformer} or LSTM~\cite{lstm} to obtain the video features. %VLMs using image encoders operate at frame-level and thus require temporal aggregation to embed the video. 
The vision and text encoders can be jointly trained using either~a)~contrastive loss %These image/video encoders can be trained jointly with text encoders using contrastive loss 
that aligns both modalities in a shared latent space e.g., CLIP~\cite{clip}, MIL-NCE~\cite{miech2020end},~b)~masked token prediction losses on the generated text~\cite{blip} aka captioning loss, or~c)~combination of captioning and contrastive losses~\cite{coca}. The vision-text features from encoders can either be fused (multimodal fusion) using attention-based mechanisms or by computing cross-modal similarity scores. 
% Alternatively, generative vision-text models use encoder-decoder architecture and are trained using masked token prediction losses on the generated text~\cite{blip}. Recent work has also proposed training VLMs using both of these losses~\cite{coca}. 
%can then be fused (multimodal fusion) together to obtain a joint representation for a downstream task using attention-based mechanisms. Instead of such explicit fusion, some VLMs also use cross-modal similarity scores for implicit fusion. 
% Some methods \cite{luo2022clip4clip,coca,violin_dataset} use a sequence model, e.g., transformers \cite{tf_transformer} or LSTM\cite{lstm}, to aggregate video features to adapt to longer videos.
We investigate 6 VLMs that span the space of these characteristics for \etv. 

%\begin{itemize}
    \noindent \textbf{CLIP4Clip~\cite{luo2022clip4clip}} uses CLIP-based \cite{clip} text and image encoders. %, followed by ViViT-like~\cite{} temporal transformer for video embedding. 
     Parameterized (e.g., LSTM-based) or non-parameterized (e.g., mean-pooling) aggregation of the resultant image features allows video representation using a single feature vector, without any explicit fusion.
    %image-text similarity computations are used for implicity fusing the modalities. 
    %The image features over time are then processed using a Transformer. They further propose parameterized e.g., LSTM-based and non-parameterized e.g., mean-pooling mechanisms for fusing \rd{unclear whether the similarity calculation is per frame or between video and text..}
    \noindent \textbf{CLIP Hitchhiker~\cite{clip_hitchiker}} uses a similar encoder structure as CLIP4Clip \cite{luo2022clip4clip} and performs weighted-mean pooling of frame embeddings using text-visual similarity scores.  %Instead of doing temporal aggregation of image representations to encode the video, it uses per-frame scoring mechanism based on image-text similarity. This is followed by a simple 
    % A weighted mean of the per-frame embeddings based on the image-text similarity scores is used to embed the video and text into a single feature vector, without any explicit fusion. %Given that the image-text similarity is implicitly used to compute this feature vector, explicit fusion of video and text input is not needed. 
    \noindent \textbf{CoCa~\cite{coca}} uses image-text (dual) encoder-decoder architecture trained using contrastive and captioning loss. The frame-level features are pooled via attentional pooling~\cite{attention_pooling} to model the temporal sequence of the video and then fused with the text features for downstream tasks.
    \noindent \textbf{MIL-NCE~\cite{miech2020end}} learns to encode video and text into a single vector using separate encoders (S3D\cite{s3d}, word2vec\cite{word2vec}) learned from scratch using the proposed multi-instance contrastive loss. 
    \noindent \textbf{VideoCLIP~\cite{videoclip}} is built on top of MIL-NCE \cite{miech2020end}. It adds additional transformer layers for video and text encoders, trained using contrastive loss. The resultant embeddings can be fused together for downstream tasks.
    \noindent \textbf{VIOLIN~\cite{violin_dataset}} uses pre-trained image/video (e.g., ResNet~\cite{resnet}, I3D~\cite{i3d}) and text encoders (e.g., GloVe~\cite{glove}, BERT~\cite{bert_model}) separately and fuses the resultant representations from each modality using bi-directional attention~\cite{bidaf}.~For each of the above VLMs, we freeze the pretrained feature extractors and finetune a fully-connected probe layer, along with the temporal aggregation layers where appropriate (CLIP4Clip-LSTM,~VIOLIN), using \etv's train split. 

    %%%%%%%%%%%%%%%%%%%%%%%%%%%%%%%%%%%%%%%%%%%%

    \subsection{Text2Text Baseline Model}
    \label{appendix:upper_bound_models}
    
    In this baseline, we generate video captions from ground-truth objects and sub-task labels and calculate a text similarity score (cosine similarity) between the video caption and the task description using a pretrained RoBERTa model~\cite{roberta} by using a manually set threshold. Videos are split into segments, with each caption containing the segment index (for temporal grounding), scene location (kitchen), (\emph{top-k}) objects in the scene, and the activity (sub-task). Refer to Table~\ref{text2text_examples} for examples. This baseline mirrors Socratic Models~\cite{socratic} which generalizes in zero-shot by leveraging the multimodal capabilities from several pretrained models. For instance, the objects from each segment can be captured using open-vocabulary VLMs~\cite{glip,regionclip,coca}, while the activities for each segment can be detected using \cite{miech2020end,videoclip} by zero-shot classification with text-to-video feature similarity. Similarly, a final summarized video caption for the whole video can be generated using an LLM. %\red{RH: missing references}
    Notable, despite having ground-truth textual representations of objects and sub-tasks on a scene-by-scene basis, the Text2text baseline model fails to generalize. We attribute this to two reasons:~(1) The pretrained RoBERTa model has limited capacity to capture (out-of-domain) word-level sub-task orderings to determine entailment in \etv, and~(2) Text2Text lacks visual inputs and might suffer from lack of inferring relationships between objects.

    % As shown in Table~\ref{table:appendix_baseline_results}, we also experiment with different combinations of visual (ResNet~\cite{resnet}, I3D~\cite{i3d}, S3D~\cite{s3d}, MViT~\cite{mvit}, CLIP~\cite{clip}) and text encoders (GloVe~\cite{glove}, BERT~\cite{bert_model}, CLIP~\cite{clip}) for the VIOLIN baseline with and without bidirectional attentional flow~\cite{bidaf}.
    % evaluate its performance on our generalization splits.
%\end{itemize}

%%%%%%%%%%%%%%%%%%%%%%%%%%%%%%%%%%%%%%%%%%%%%%%%%%%%%

\begin{figure*}[t]
    \begin{tcolorbox}[fontupper=\small, fontlower=\Large, title=Text2Text Baseline Examples]
        % \begin{multicols}{2}
            \noindent Text Description: \code{apple is cooled in a Fridge and cleaned in a SinkBasin}

            \noindent Video Caption:\\
            \code{Segment: 1. Location: kitchen. Objects: countertop, apple. Activity: go to countertop}\\
            \code{Segment: 2. Location: kitchen. Objects: fridge, apple. Activity: go to fridge}\\
            \code{Segment: 3. Location: kitchen. Objects: apple, fridge. Activity: cool apple}\\
            \code{Segment: 4. Location: kitchen. Objects: apple, fridge. Activity: cool apple}\\
            \code{Segment: 5. Location: kitchen. Objects: sink, apple. Activity: go to sink}\\
            \code{Segment: 6. Location: kitchen. Objects: apple, sink. Activity: clean apple} 

            \noindent Task Verified: \answer{True}
            \\

            \noindent Text Description: \code{lettuce is picked, cooled in a Fridge, and sliced in a SinkBasin}

            \noindent Video Caption:\\
            \code{Segment: 1. Location: kitchen. Objects: sink, lettuce. Activity: go to sink}\\
            \code{Segment: 2. Location: kitchen. Objects: sink, lettuce. Activity: go to sink}\\
            \code{Segment: 3. Location: kitchen. Objects: lettuce, sink. Activity: clean lettuce}\\
            \code{Segment: 4. Location: kitchen. Objects: fridge, lettuce. Activity: go to fridge}\\
            \code{Segment: 5. Location: kitchen. Objects: lettuce, fridge. Activity: cool lettuce} 

            \noindent Task Verified: \code{\red{False}}
        % \end{multicols}
    \end{tcolorbox}
    \caption{Text2Text Baseline Examples.}
    \label{text2text_examples}
\end{figure*}

%%%%%%%%%%%%%%%%%%%%%%%%%%%%%%%%%%%%%%%%%%%%%%%%    

\input{tables/sup_list_of_tasks}
\input{figures/sup/extra-analysis}

%% file: tables/sup_list_of_tasks.tex
\begin{table*}[ht]
\centering
\begin{adjustbox}{width=\textwidth}
\begin{tabular}{|l|l|l|l|}
\hline
\textbf{Complex}\textbackslash \textbf{Order} & \textbf{0} & \textbf{1} & \textbf{2} \\ \hline
\textbf{1} & \begin{tabular}[c]{@{}l@{}}
clean\_simple, cool\_simple, heat\_simple\\
pick\_simple, place\_simple, slice\_simple\end{tabular} &  &  \\ \hline
\textbf{2} & \begin{tabular}[c]{@{}l@{}}
\blue{clean\_and\_cool}, clean\_and\_heat\\ clean\_and\_place, clean\_and\_slice\\
cool\_and\_place, heat\_and\_place\\
slice\_and\_cool, slice\_and\_heat\\ slice\_and\_place\end{tabular} & 
\begin{tabular}[c]{@{}l@{}}
\blue{clean\_then\_cool}, clean\_then\_heat\\ clean\_then\_place, clean\_then\_slice\\ \blue{cool\_then\_clean}, cool\_then\_place\\ cool\_then\_slice, heat\_then\_clean\\ 
heat\_then\_place, heat\_then\_slice\\
slice\_then\_clean, slice\_then\_cool\\ 
slice\_then\_heat, slice\_then\_place\end{tabular} &  \\ \hline
\textbf{3} & \begin{tabular}[c]{@{}l@{}}slice\_and\_clean\_and\_place,
\blue{cool\_and\_clean\_and\_place}\\ cool\_and\_slice\_and\_place, 
heat\_and\_clean\_and\_place\\ \blue{slice\_and\_heat\_and\_place},
slice\_and\_heat\_and\_clean\\ \blue{cool\_and\_slice\_and\_clean}\end{tabular} & 

\begin{tabular}[c]{@{}l@{}}\end{tabular} & \begin{tabular}[c]{@{}l@{}}\blue{clean\_then\_cool\_then\_place},
\blue{clean\_then\_cool\_then\_slice}\\ clean\_then\_heat\_then\_place,
clean\_then\_heat\_then\_slice\\ \blue{clean\_then\_slice\_then\_cool},
clean\_then\_slice\_then\_heat\\ \blue{cool\_then\_clean\_then\_place},
cool\_then\_clean\_then\_slice\\ \blue{cool\_then\_slice\_then\_clean},
heat\_then\_clean\_then\_place\\ heat\_then\_clean\_then\_slice,
heat\_then\_slice\_then\_clean\\ \blue{slice\_then\_clean\_then\_cool}, slice\_then\_clean\_then\_heat\\ slice\_then\_clean\_then\_place, \blue{slice\_then\_cool\_then\_clean}\\ slice\_then\_cool\_then\_place, slice\_then\_heat\_then\_clean\\ \blue{slice\_then\_heat\_then\_place},
\blue{clean\_and\_cool\_then\_place}\\ \blue{clean\_and\_cool\_then\_slice}, clean\_and\_heat\_then\_place\\ clean\_and\_heat\_then\_slice, \blue{clean\_and\_slice\_then\_cool}\\ clean\_and\_slice\_then\_heat,
\blue{clean\_then\_cool\_and\_slice}\\ clean\_then\_heat\_and\_slice,
\blue{cool\_and\_slice\_then\_clean}\\ \blue{cool\_then\_clean\_and\_slice}, heat\_and\_slice\_then\_clean\\ heat\_then\_clean\_and\_slice, slice\_and\_clean\_then\_place\\ slice\_and\_cool\_then\_place, \blue{slice\_and\_heat\_then\_place}\\ \blue{slice\_then\_clean\_and\_cool}, slice\_then\_clean\_and\_heat\\ \blue{clean\_then\_cool\_and\_place}, clean\_then\_heat\_and\_place\\ clean\_then\_slice\_and\_place,
slice\_then\_cool\_and\_place\\ \blue{slice\_then\_heat\_and\_place}, slice\_then\_clean\_and\_place\\ heat\_then\_clean\_and\_place, \blue{heat\_then\_slice\_and\_place}\\ \blue{cool\_then\_clean\_and\_place}, cool\_then\_slice\_and\_place\end{tabular} \\ \hline
\end{tabular}
\end{adjustbox}
\vspace{5pt}
\caption{\textbf{List of tasks in \etv dataset} arranged according to complexity (rows) and ordering (columns) in the tasks. A total of 82 tasks were considered for the dataset generation, split into train and novel composition sets. \blue{Blue} denotes novel composition tasks.}
\label{tab:list_of_tasks}
\end{table*}

%% file: figures/sup/extra-analysis.tex
\begin{figure*}
\centering
    \includegraphics[width=\linewidth]{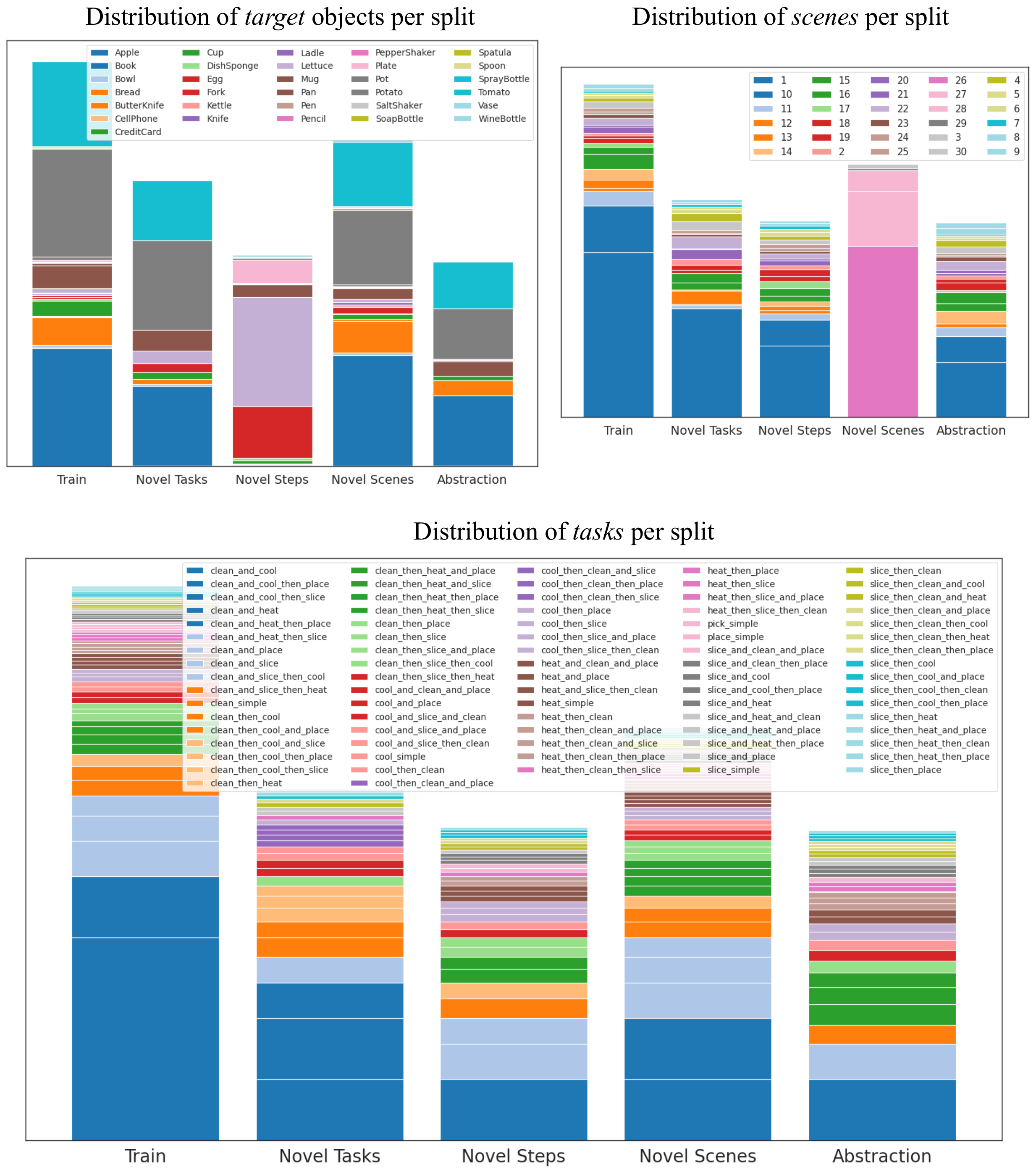}
    \caption{\etv dataset analysis. Distribution of \emph{target} objects [row 1, left], \emph{kitchen-scenes} [row 1, right], and \emph{sub-tasks} [row 2] in each split. The Y-axis is in the log scale.}
    \label{figure:tasks-all-splits}
\end{figure*}